\documentclass{article}

\PassOptionsToPackage{numbers, compress}{natbib}
\usepackage[main, final]{neurips_2026}

\usepackage[utf8]{inputenc} 
\usepackage[T1]{fontenc}    
\usepackage{hyperref}       
\usepackage{url}            
\usepackage{booktabs}       
\usepackage{microtype}      

\usepackage{multirow}
\usepackage{amsmath}
\usepackage{amssymb}
\usepackage{graphicx}
\usepackage{placeins}
\usepackage[table]{xcolor}  
\usepackage{caption}
\usepackage{fontawesome5}

\definecolor{oursrow}{HTML}{EFF5FB}
\definecolor{sectionlabel}{HTML}{666666}

\title{The Mask Is Not the Object: Volumetric Supervision for 3D Gaussian Segmentation}

\author{
Jeonghwan Cho
  \hspace{1.4em}
Minsu Kim
  \hspace{1.4em}
Junyoung Hong
  \hspace{1.4em}
Seon Joo Kim \\[1.0em] Yonsei University
}

\begin{document}

\maketitle
\vspace{-2em}

{\centering\small
\href{https://jhxcho.github.io/voda/}{%
    {\faGlobe\enspace\textbf{Project Page}}%
}\par}

\vspace{1.2em}

\raggedbottom

\begin{abstract}
3D Gaussian Splatting enables object-level operations on Gaussian primitives, yet existing segmentation methods are largely trained and evaluated through rendered views.
This mismatch can produce accurate rendered masks from incomplete 3D selections since alpha-blended feature supervision concentrates learning on dominant contributors while leaving occluded or low-contribution Gaussians weakly supervised.
We propose VODA, a plug-and-play supervision objective that exploits local depth structure to group per-pixel contributors and directly supervises all Gaussians in selected depth clusters, bypassing contribution-weighted gradient delivery.
To evaluate volumetric completeness, we introduce Removal-3D, a synthetic object-removal benchmark, and the Removal Depth Error (RDE), a metric comparing predicted and ground-truth depth changes.
Across four rendering-based baselines, VODA consistently improves volumetric segmentation on Removal-3D while maintaining or improving performance on LERF and ScanNet.
Further analysis shows that VODA recovers lower-contribution Gaussians, while simple gradient reweighting and spatial propagation fail to reproduce comparable gains.
\end{abstract}

\suppressfloats[t]
\begin{figure}[t]
\centering
\includegraphics[trim={0 0 0 60pt}, width=\linewidth]{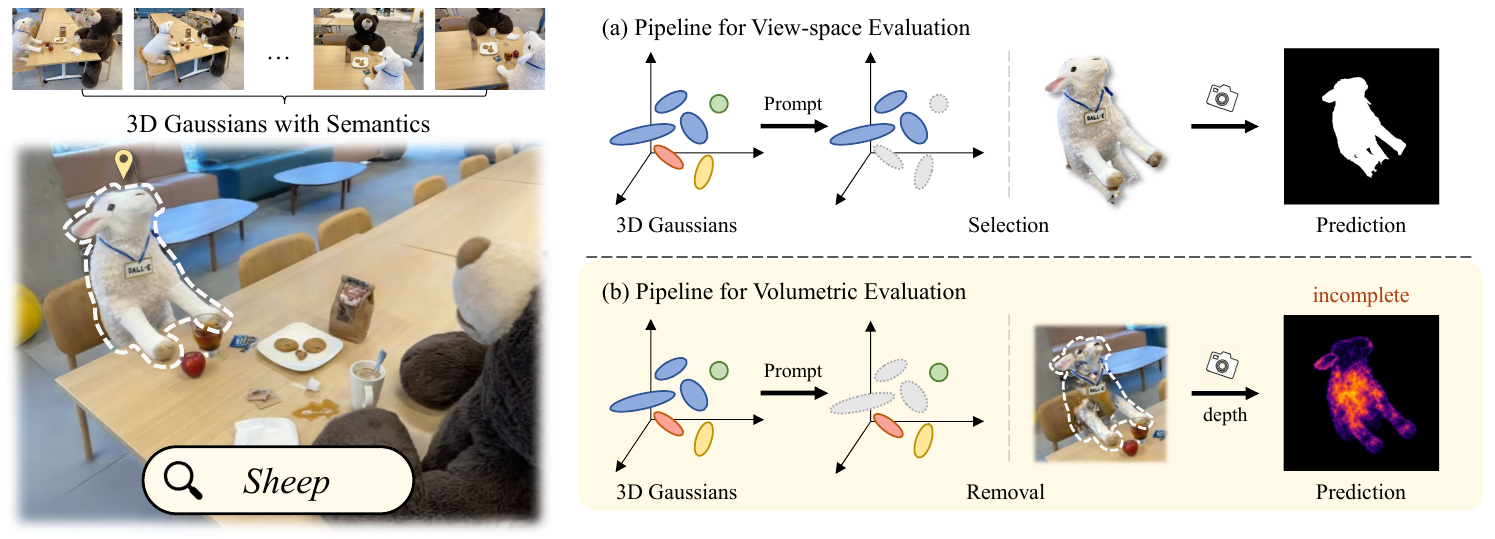}
\caption{
\textbf{A plausible rendered mask does not imply complete 3D selection.} For the query ``sheep,'' (a) the selected Gaussians render a plausible object mask, but (b) the depth change after removing the same Gaussian set indicates incomplete object removal in the reconstructed scene.
This exposes a gap between rendered-mask quality and Gaussian-level selection completeness.
}
\label{fig:motivation}
\end{figure}

\section{Introduction}
\label{sec:introduction}

Object-level editing of reconstructed scenes requires selecting objects as complete 3D entities.
Removing, relocating, or extracting an object depends on capturing its full spatial extent, including its interior.
For scenes represented by 3D Gaussian Splatting (3DGS)~\cite{kerbl20233dgs}, this means selecting all Gaussians associated with the object.
Yet existing segmentation methods~\cite{qin2024langsplat, shi2024legaussians, ye2024gaussiangrouping, jsmbankILGS, wu2024opengaussian, cen2025laga} are commonly evaluated by the accuracy of their rendered 2D masks, which does not establish whether the underlying 3D selection is complete.
Figure~\ref{fig:motivation} illustrates this gap: a selection can produce an accurate mask yet miss some of the object's Gaussians, leaving visible remnants after removal.

This gap stems from rendering-based feature supervision.
Existing methods~\cite{ye2024gaussiangrouping, jsmbankILGS, wu2024opengaussian, cen2025laga} learn per-Gaussian instance features through rendered 2D feature maps, using external feature or mask targets, or mask-based contrastive objectives.
Despite their differences, these objectives deliver gradients according to each Gaussian's rendering contribution (\S\ref{sec:prelim_feature_learning}).
Dominant contributors can thus suffice to render an accurate mask, while low-contribution Gaussians belonging to the same object receive weak or no supervision and may remain unselected.

We therefore study \emph{volumetric 3D Gaussian segmentation}, aiming to select object-associated Gaussians beyond those needed to render a correct mask.
We propose \textbf{VODA}, a plug-and-play objective that extends object-level supervision beyond the dominant contributors.
VODA modifies only feature supervision, leaving the baseline's reconstructed scene, feature representation, and inference-time selection procedure unchanged.

Our key observation is that, after RGB optimization, Gaussians associated with the same object tend to cluster at similar depths, while larger depth gaps often reflect separation between distinct objects.
VODA exploits this structure by grouping per-pixel contributors into depth clusters and selecting the cluster that best explains each pixel (\S\ref{sec:depth_aware_rasterizer}).
Targets constructed from rendered features are then applied directly to every Gaussian in the selected cluster, supervising both dominant and low-contribution members while bypassing the contribution-weighted gradient path (\S\ref{sec:direct_gaussian_supervision}).
This direct supervision extends coverage within objects, while a complementary pixel-level push loss sharpens boundaries between adjacent objects.

Evaluating volumetric segmentation requires measuring whether an object selection is complete beyond its visible projection.
Existing benchmarks score rendered masks or visible surface labels and therefore do not directly measure this property.
We use object removal as a diagnostic: removing a complete selection should produce the same depth change as removing the target object itself.
We therefore introduce Removal-3D, a synthetic indoor-scene benchmark with paired ground-truth depth maps before and after target removal.
To quantify how closely a predicted selection reproduces this change, we propose Removal Depth Error (RDE), which compares the ground-truth depth change with the change induced by removing the predicted Gaussians.

Experiments across four rendering-based baselines~\cite{wu2024opengaussian, cen2025laga, ye2024gaussiangrouping, jsmbankILGS} show that VODA consistently improves volumetric segmentation on Removal-3D while maintaining or improving performance on standard benchmarks~\cite{kerr2023lerf, dai2017scannet}.
Ablations identify direct Gaussian-level supervision as the primary source of these gains, while additional diagnostics support the need to change which Gaussians receive supervision rather than merely modifying the existing rendering-based signal.

In summary, our contributions are threefold.
(1) We formulate volumetric 3D Gaussian segmentation, distinguishing selection completeness from rendered-mask accuracy and identifying the bias in rendering-based supervision.
(2) We propose VODA, a plug-and-play objective that directly supervises all Gaussians in selected depth clusters.
(3) We introduce Removal-3D and RDE, a benchmark and metric that assess selection completeness through removal-induced depth changes.

\section{Related Work}
\label{sec:related_work}

The explicit primitives of 3D Gaussian Splatting (3DGS)~\cite{kerbl20233dgs} support object-level selection and editing.
Recent methods attach semantic features or object identities to Gaussians for object-level scene understanding~\cite{ye2024gaussiangrouping, zhu2025objectgs}.
We focus on how 2D supervision is transferred to Gaussians, distinguishing rendering-based feature learning from direct 2D-to-Gaussian association.

\begin{figure}
\centering
\includegraphics[width=\linewidth]{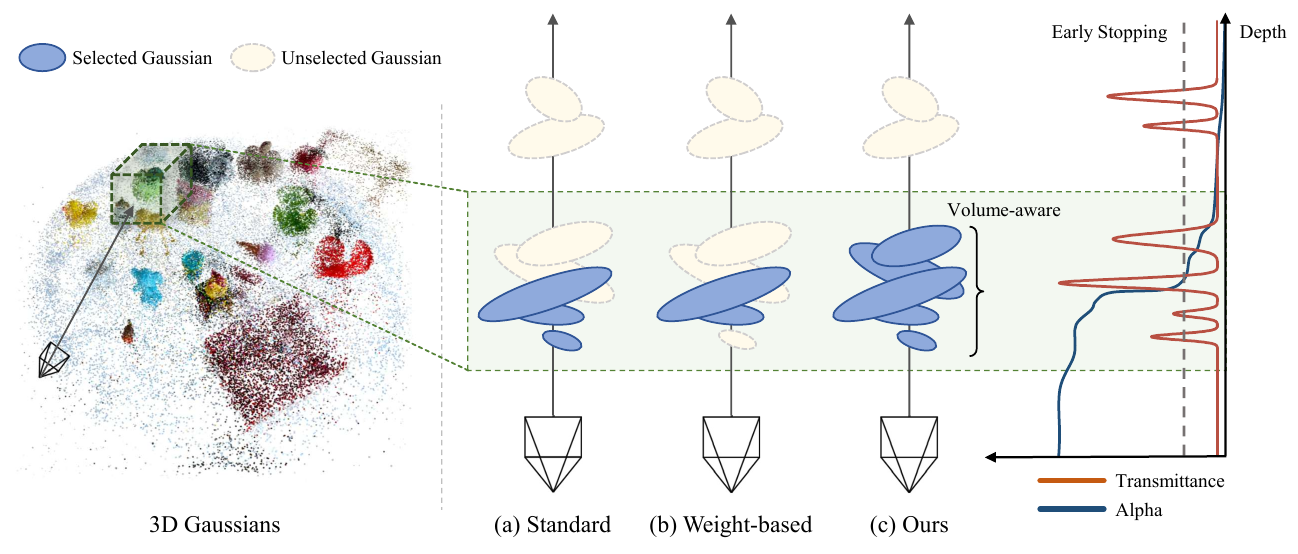}
\caption{
\textbf{Comparison of supervision and association strategies for per-pixel Gaussian contributors.} (a) Rendering-based feature learning delivers contribution-weighted gradients, giving weak supervision to low-contribution Gaussians.
(b) Direct association links each pixel to a sparse subset of contributors selected by contribution-based rules.
(c) VODA directly supervises every Gaussian in the selected depth cluster, including low-contribution Gaussians associated with the same object.
}
\label{fig:rasterizer_comparison}
\end{figure}

\paragraph{Rendering-based Feature Learning.}
\label{sec:rw_rendering_based_feature_learning}
Earlier NeRF methods lift 2D semantics through feature distillation, mask supervision, or contrastive learning~\cite{zhi2021semanticnerf, kobayashi2022dff, tschernezki2022n3f, siddiqui2023panopticlifting, bhalgat2023contrastivelift, liu20233dovs, cen2023sa3d, kim2024garfield, ying2024omniseg3d, bhalgat2024n2f2}, including distillation of pretrained visual features such as DINO~\cite{caron2021dino}.
Rendering-based 3DGS methods similarly alpha-blend per-Gaussian features into 2D maps, with two main supervision paradigms.
\textbf{Supervised feature learning} matches these maps to external targets: CLIP features~\cite{qin2024langsplat, shi2024legaussians, zhou2024feature3dgs, zuo2024fmgs, qu2024goi, li2024langsurf}, SAM mask identities as in Gaussian Grouping~\cite{ye2024gaussiangrouping}, or both jointly~\cite{jsmbankILGS}.
\textbf{Self-supervised feature learning} instead uses SAM masks as grouping cues, pulling features within a mask together and pushing features across masks apart without explicit feature targets~\cite{wu2024opengaussian, cen2025laga, cen2025saga, choi2024clickgaussian, lee2025open3drf}.
Although their supervision sources differ, these rendering-based objectives share the contribution-weighted gradient path analyzed in \S\ref{sec:prelim_feature_learning}.
As illustrated in Figure~\ref{fig:rasterizer_comparison}a, low-contribution Gaussians can therefore receive weak supervision even when the rendered features match the 2D targets or grouping constraints.
VODA complements either paradigm with direct Gaussian-level supervision while leaving the reconstructed scene, feature representation, and inference-time selection unchanged.

\paragraph{Direct 2D-to-Gaussian Association.}
\label{sec:rw_direct_association}
Another line of work associates 2D observations directly with Gaussians rather than learning features through rendered-map losses.
Lifting By Gaussians~\cite{chacko2025lbg} assigns 2D mask labels to the per-pixel maximum-contributing Gaussian and merges the resulting fragments across views.
Dr.~Splat~\cite{kim2025drsplat} registers CLIP embeddings to dominant or Top-$k$ Gaussians for optimization-free 3D search.
Quantile Rendering~\cite{jeong2025qrender} uses quantile-based sampling to associate high-dimensional features with a sparse set of influential Gaussians.
Despite their different association rules, these methods associate each pixel with only a sparse subset of Gaussian contributors, as illustrated in Figure~\ref{fig:rasterizer_comparison}b.
Multi-view observations can expand this set, but object-associated Gaussians that remain weak contributors across views may still receive no assignment.
Thus, bypassing rendered-feature optimization alone does not necessarily resolve the incompleteness induced by sparse, contribution-based association.
VODA instead uses per-pixel depth structure to identify the target-associated cluster and directly supervises all Gaussians within it.
\section{Method}
\label{sec:method}

We propose VODA, a plug-and-play supervision objective for volumetric 3D Gaussian segmentation.
We first analyze why standard rendering-based supervision is insufficient in \S\ref{sec:prelim_feature_learning}.
The depth-aware rasterizer in \S\ref{sec:depth_aware_rasterizer} then groups per-pixel contributors and selects a target depth cluster.
Direct Gaussian-level supervision in \S\ref{sec:direct_gaussian_supervision} applies a feature loss to every Gaussian in the selected cluster, together with a complementary pixel-level push loss that encourages separation between object instances.

\subsection{Contribution-weighted Supervision in Feature Rendering}
\label{sec:prelim_feature_learning}

\begin{figure*}
\centering
\includegraphics[width=\linewidth]{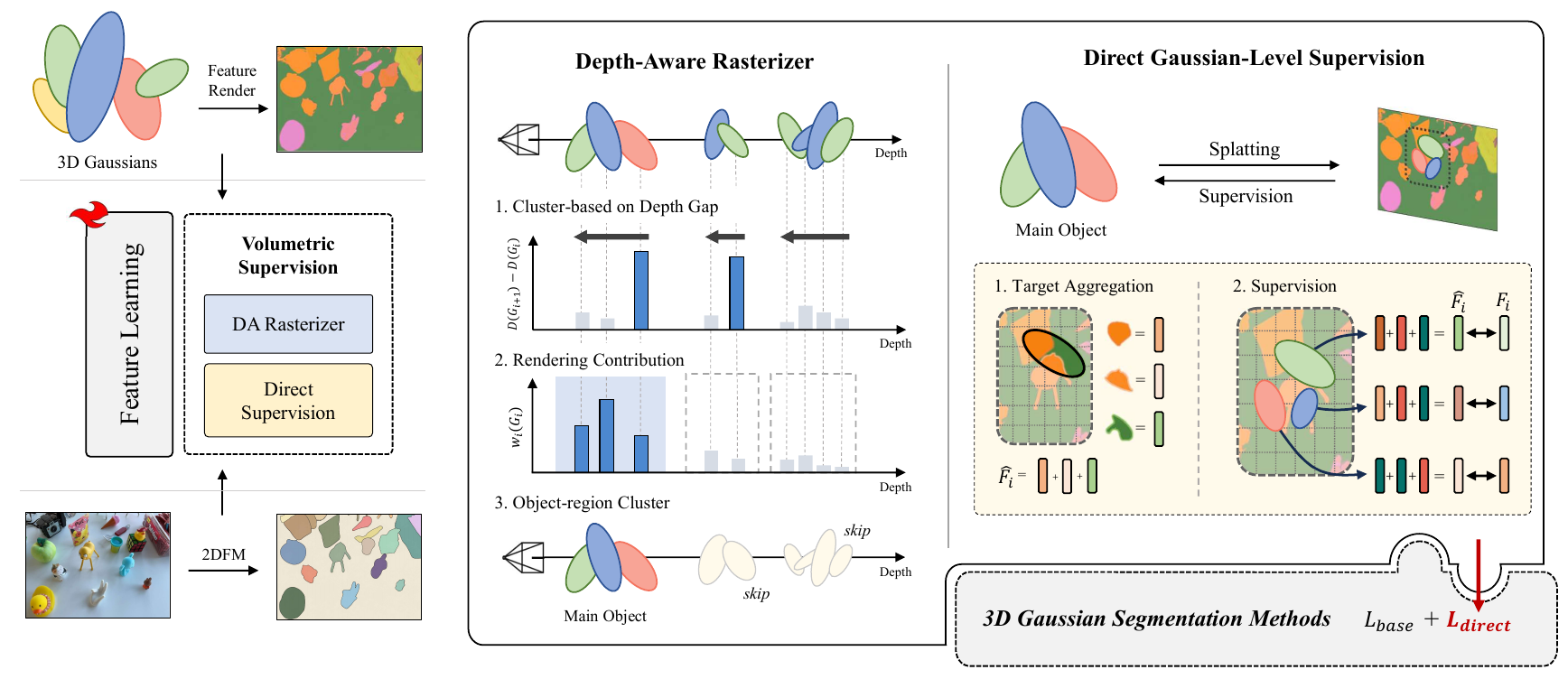}
\caption{
\textbf{Overview of VODA.} The depth-aware rasterizer groups per-pixel contributors using adjacent depth gaps and selects the cluster with the largest total rendering contribution.
Alpha-weighted pixel features are aggregated into detached Gaussian-level targets.
A direct feature loss supervises every Gaussian in the selected cluster, including low-contribution Gaussians, without contribution-weighted gradient delivery.
The baseline feature-rendering loss remains active.
}
\label{fig:pipeline}
\end{figure*}

Let $\{g_i\}_{i=1}^{N}$ be the Gaussians of a reconstructed 3DGS scene, and let $\mathbf{f}_i\in\mathbb{R}^{C}$ be the learnable instance feature associated with Gaussian $g_i$.
For a pixel $p$, the Gaussians overlapping the pixel are sorted by camera-space depth and alpha-blended into a rendered feature map:
\begin{equation}
    \mathbf{F}(p)=\sum_{i\in\mathcal{G}(p)} w_i(p)\mathbf{f}_i,
    \qquad
    w_i(p)=T_i(p)\alpha_i(p),
    \label{eq:feature_rendering}
\end{equation}
where $\mathcal{G}(p)$ denotes the Gaussians contributing to pixel $p$ in depth order, $\alpha_i(p)$ is the per-pixel alpha value of $g_i$, and $T_i(p)$ is the accumulated transmittance before $g_i$ in that ordering.

Let $\mathcal{L}_{\mathrm{rend}}$ denote the component of the baseline feature-learning objective that is computed through the rendered feature maps.
For this rendered-map supervision term, applying the chain rule gives:
\begin{equation}
    \frac{\partial\mathcal{L}_{\mathrm{rend}}}{\partial \mathbf{f}_i}
    =
    \sum_{p:i\in\mathcal{G}(p)}
    w_i(p)
    \frac{\partial\mathcal{L}_{\mathrm{rend}}}{\partial \mathbf{F}(p)} .
    \label{eq:contribution_gradient}
\end{equation}
This contribution-weighted path is shared by both paradigms described in \S\ref{sec:rw_rendering_based_feature_learning}, whether rendered features are matched to external targets or compared across pixels.
As Eq.~\ref{eq:contribution_gradient} shows, this shared path leads to an imbalance in gradient delivery: dominant contributors can learn features sufficient for an accurate rendered mask, while low-contribution Gaussians associated with the same object receive weak gradients due to their small rendering weights.

\subsection{Depth-Aware Rasterizer}
\label{sec:depth_aware_rasterizer}

To address this imbalance, we need to identify object-associated Gaussians beyond those retained by maximum or Top-$k$ selection (Figure~\ref{fig:rasterizer_comparison}).
Instead of relying solely on individual rendering contributions, we exploit the local depth structure of the reconstructed scene.
During reconstruction, cloning and splitting introduce new Gaussians around existing primitives~\cite{kerbl20233dgs}.
This local refinement motivates using depth proximity as a cue for object association: contributors representing the same local object region can form a depth cluster containing both dominant and low-contribution Gaussians.
Larger gaps can indicate separation from other objects or the background.

\paragraph{Depth clustering.}
To group contributors based on these depth gaps, we introduce a depth-aware rasterizer that collects per-pixel contributors, their camera-space depths, and alpha values.
It disables transmittance-based early termination to retain contributors behind dominant Gaussians.
For each pixel $p$ with at least two contributors, let $d_j(p)$ denote the camera-space depth of the $j$-th contributor in depth order.
We split consecutive contributors whenever their depth gap exceeds $\tau(p)$:
\begin{equation}
    \tau(p)=
    \mathrm{clip}
    \left(
        \gamma\cdot
        \mathrm{median}_{1\leq j<|\mathcal{G}(p)|}
        \left[d_{j+1}(p)-d_j(p)\right],
        \tau_{\min},
        \tau_{\max}
    \right).
    \label{eq:depth_threshold}
\end{equation}
The median adapts the threshold to the local gap distribution, $\gamma$ controls split sensitivity, and clipping bounds it within $[\tau_{\min},\tau_{\max}]$.
We denote the resulting depth clusters by $\mathcal{C}_k(p)$.

\paragraph{Target cluster selection.}
To identify the target cluster, VODA selects the cluster with the largest total contribution at each pixel:
\begin{equation}
    \mathcal{C}^{*}(p)
    =
    \arg\max_{\mathcal{C}_k(p)}
    \sum_{i\in\mathcal{C}_k(p)} w_i(p).
    \label{eq:selected_cluster}
\end{equation}
Using total contribution makes cluster selection less sensitive to near-camera floater Gaussians than relying on proximity alone.

\subsection{Direct Gaussian-Level Supervision}
\label{sec:direct_gaussian_supervision}

Given the selected depth cluster, we directly supervise its Gaussian features to bypass the contribution-weighted rendering path.
Since there are no ground-truth instance features for individual Gaussians, we construct Gaussian-level pseudo-targets from the full rendered feature $\mathbf{F}(p)$ defined in Eq.~\ref{eq:feature_rendering}, rather than rendering the selected cluster alone.
This feature is dominated by high-contribution Gaussians that receive strong baseline supervision, providing a learned feature signal for other Gaussians in the same cluster.
We detach the resulting Gaussian-level targets with $\mathrm{sg}[\cdot]$ and apply the loss directly to Gaussian features, while retaining the baseline's original rendering-based supervision.

\paragraph{Gaussian-level target aggregation.}
A Gaussian may be selected at multiple pixels, which we collect as $\mathcal{P}_i=\{p\mid g_i\in\mathcal{C}^{*}(p)\}$.
This set contains only pixels where $g_i$ is selected, not every pixel to which it contributes.
VODA aggregates the rendered features over $\mathcal{P}_i$ to obtain $\hat{\mathbf{F}}_i$:
\begin{equation}
    \hat{\mathbf{F}}_i
    =
    \mathrm{sg}\!\left[
    \frac{\sum_{p\in\mathcal{P}_i}\alpha_i(p)\mathbf{F}(p)}
         {\sum_{p\in\mathcal{P}_i}\alpha_i(p)+\epsilon}
    \right],
    \label{eq:target_aggregation}
\end{equation}
where $\epsilon$ ensures numerical stability and $\alpha_i(p)$ weights each target by the Gaussian's alpha at $p$.

\paragraph{Direct feature loss.}
The direct loss aligns selected Gaussian features with aggregated targets:
\begin{equation}
    \mathcal{L}_{\mathrm{direct}}
    =
    \frac{1}{|\mathcal{G}_{\mathrm{sel}}|}
    \sum_{g_i\in\mathcal{G}_{\mathrm{sel}}}
    \left\|\mathbf{f}_i-\hat{\mathbf{F}}_i\right\|_2^2,
    \qquad
    \mathcal{G}_{\mathrm{sel}}=\{g_i\mid |\mathcal{P}_i|>0\}.
    \label{eq:dir_loss}
\end{equation}
The loss acts directly on Gaussian features with detached targets, so its gradient bypasses rendering weights and reaches low-contribution Gaussians in the selected clusters.
Eq.~\ref{eq:dir_loss} uses squared-$\ell_2$ by default, while baseline-specific distance choices are provided in the supplementary.

\paragraph{Pixel-level feature push.}
A depth cluster can span adjacent object instances, so we add a pixel-level push loss to reduce feature mixing across object boundaries.
Within each training view, we randomly sample a set $\mathcal{Q}$ of pixel pairs $(p,q)$ belonging to different non-background masks and penalize pairs whose rendered-feature cosine similarity exceeds a margin $m$ in the learned feature space:
\begin{equation}
    \mathcal{L}_{\mathrm{push}}
    =
    \frac{1}{|\mathcal{Q}|}
    \sum_{(p,q)\in\mathcal{Q}}
    \max\!\left(0,\,
    \cos\!\left(\mathbf{F}(p),\mathbf{F}(q)\right) - m
    \right).
    \label{eq:push_loss}
\end{equation}

\subsection{Training Objective}
\label{sec:training_objective}

Let $\mathcal{L}_{\mathrm{base}}$ denote the baseline feature-learning objective, including any existing Gaussian-level regularization.
We retain this objective and add VODA's direct and pixel-level push losses:
\begin{equation}
    \mathcal{L}_{\mathrm{VODA}}
    =
    \lambda_{\mathrm{direct}}\mathcal{L}_{\mathrm{direct}}
    +
    \lambda_{\mathrm{push}}\mathcal{L}_{\mathrm{push}},
    \qquad
    \mathcal{L}
    =
    \mathcal{L}_{\mathrm{base}}
    +
    \mathcal{L}_{\mathrm{VODA}}.
    \label{eq:total_loss}
\end{equation}
Depending on the baseline, VODA is applied during feature learning on a reconstructed scene or in a post-training feature-only fine-tuning stage.
Scene geometry, opacity, feature representation, and inference-time selection remain unchanged, and VODA adds no parameters or computation at inference.
Implementation schedules and hyperparameters are provided in the supplementary.
\section{Experiments}
\label{sec:experiments}

We evaluate VODA on two standard 3DGS segmentation benchmarks, LERF and ScanNet, in \S\ref{sec:exp_compat}, and on our Removal-3D benchmark for volumetric evaluation in \S\ref{sec:exp_removal}.
We further ablate its design choices in \S\ref{sec:exp_ablation} and analyze recovered Gaussians and contribution-based alternatives in \S\ref{sec:exp_analysis}.
We evaluate four baselines spanning both feature-learning paradigms in \S\ref{sec:rw_rendering_based_feature_learning}: OpenGaussian~\cite{wu2024opengaussian} and LaGa~\cite{cen2025laga} are self-supervised, while Gaussian Grouping~\cite{ye2024gaussiangrouping} and ILGS~\cite{jsmbankILGS} are supervised.

\paragraph{Implementation details.}
We implement VODA on the official codebase of each baseline and apply it only to the feature-learning stage, leaving RGB reconstruction and inference unchanged.
The depth-clustering and loss hyperparameters are shared across all four baselines.
VODA is applied within the existing feature stage for OpenGaussian and LaGa, and as a short feature-only fine-tuning stage for Gaussian Grouping and ILGS.
Full implementation details are provided in the supplementary.

\begin{table}[t]
\centering
\caption{Open-vocabulary 3D object selection on LERF under the select-then-render protocol.
We report rendered-mask mIoU and mAcc@0.25.
\texttt{waldo\_kitchen} is abbreviated as \texttt{kitchen}.}
\label{table:lerf}
\begingroup
\centering
\begingroup
\footnotesize
\renewcommand{\arraystretch}{1.1}
\setlength{\tabcolsep}{0.8pt}
\begin{tabular}{>{\raggedright\arraybackslash\scriptsize}m{0.92in}*{10}{>{\centering\arraybackslash}m{0.433in}}}
\toprule
            & \multicolumn{5}{c}{mIoU $\uparrow$} & \multicolumn{5}{c}{mAcc. $\uparrow$} \\
\cmidrule(lr){2-6} \cmidrule(lr){7-11}
\multirow{-2}{*}{Methods} &
  \scriptsize figurines &
  \scriptsize teatime & \scriptsize ramen & \scriptsize kitchen & \textbf{Mean} &
  \scriptsize figurines &
  \scriptsize teatime & \scriptsize ramen & \scriptsize kitchen & \textbf{Mean} \\
\midrule
LangSplat~\cite{qin2024langsplat}
  & 10.16 & 11.38 & 7.92 & 9.18 & 9.66
  & 8.93 & 20.34 & 11.27 & 9.09 & 12.41 \\
LEGaussians~\cite{shi2024legaussians}
  & 17.99 & 19.27 & 15.79 & 11.78 & 16.21
  & 23.21 & 27.12 & 26.76 & 18.18 & 23.82 \\
\midrule
OpenGaussian~\cite{wu2024opengaussian}
  & 49.19 & 45.63 & 22.34 & \textbf{31.33} & 37.12
  & \textbf{69.64} & 74.58 & 32.39 & \textbf{50.00} & 56.65 \\
\rowcolor{oursrow}
\quad + VODA
  & \textbf{55.31} & \textbf{62.89} & \textbf{25.26} & 28.47 & \textbf{42.98}
  & \textbf{69.64} & \textbf{79.66} & \textbf{36.62} & 45.45 & \textbf{57.84} \\
LaGa~\cite{cen2025laga}
  & \textbf{65.79} & 60.59 & \textbf{53.35} & 47.47 & 56.80
  & \textbf{89.29} & 83.05 & 64.79 & 68.18 & 76.33 \\
\rowcolor{oursrow}
\quad + VODA
  & 63.28 & \textbf{69.05} & 50.42 & \textbf{58.38} & \textbf{60.28}
  & \textbf{89.29} & \textbf{91.53} & \textbf{67.61} & \textbf{81.82} & \textbf{82.56} \\
\midrule
Gaussian Grouping~\cite{ye2024gaussiangrouping}
  & 26.04 & 26.10 & 16.10 & \textemdash & 22.75
  & 46.43 & 46.67 & 33.33 & \textemdash & 42.14 \\
\rowcolor{oursrow}
\quad + VODA
  & \textbf{51.45} & \textbf{43.64} & \textbf{36.09} & \textemdash & \textbf{43.73}
  & \textbf{82.14} & \textbf{80.00} & \textbf{66.67} & \textemdash & \textbf{76.27} \\
ILGS~\cite{jsmbankILGS}
  & 24.43 & 38.70 & 26.87 & \textemdash & 30.00
  & 50.00 & 75.00 & 55.56 & \textemdash & 60.19 \\
\rowcolor{oursrow}
\quad + VODA
  & \textbf{42.36} & \textbf{58.32} & \textbf{48.19} & \textemdash & \textbf{49.62}
  & \textbf{75.00} & \textbf{91.67} & \textbf{77.78} & \textemdash & \textbf{81.48} \\
\bottomrule
\end{tabular}
\endgroup
\endgroup
\par\medskip
\includegraphics[trim={0pt 5pt 0pt 0pt},clip,width=\linewidth]{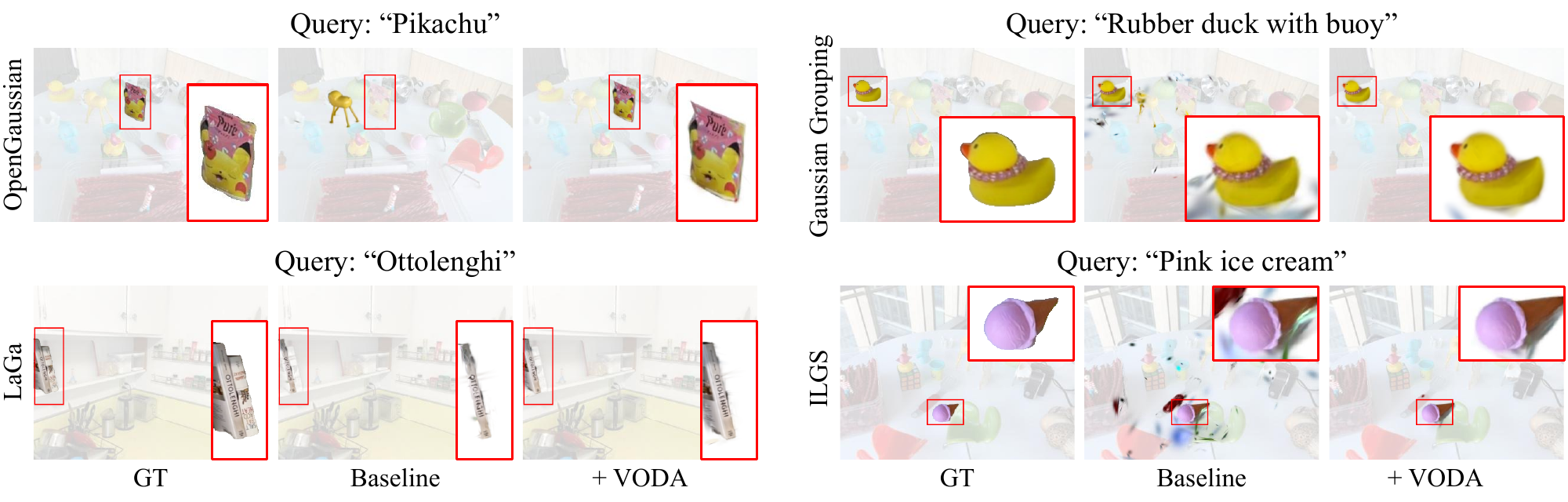}
\captionof{figure}{
Qualitative 3D object selection on LERF.
For ``Ottolenghi,'' VODA recovers object regions missed by the baselines, while for ``Pink ice cream,'' it suppresses spurious selections.
These examples demonstrate improvements in both the completeness and precision of 3D Gaussian selection.
}
\label{fig:lerf_text_query}
\end{table}

\begin{table}[t]
\centering
\caption{Text-query 3D semantic segmentation on ScanNet~\cite{dai2017scannet}.
We report point-cloud mIoU and mAcc for three category splits.
Gaussian Grouping and ILGS are omitted (see text).}
\label{table:scannet}
\begingroup
\centering
\fontsize{9pt}{10.8pt}\selectfont
\begin{tabular}{lcccccc}
\toprule
\multirow{2}{*}{Methods} & \multicolumn{2}{c}{19 classes} & \multicolumn{2}{c}{15 classes} & \multicolumn{2}{c}{10 classes} \\
\cmidrule(lr){2-3} \cmidrule(lr){4-5} \cmidrule(lr){6-7}
             & mIoU $\uparrow$ & mAcc. $\uparrow$ & mIoU $\uparrow$ & mAcc. $\uparrow$ & mIoU $\uparrow$ & mAcc. $\uparrow$ \\
\midrule
LangSplat~\cite{qin2024langsplat}    & 3.78  & 9.11  & 5.35  & 13.20 & 8.40  & 22.06 \\
LEGaussians~\cite{shi2024legaussians} & 3.84 & 10.87 & 9.01  & 22.22 & 12.82 & 28.62 \\
\midrule
OpenGaussian~\cite{wu2024opengaussian}
  & 26.20 & 39.35 & 28.00 & 42.67 & 34.57 & 48.38 \\
\rowcolor{oursrow}
\quad + VODA
  & \textbf{29.00} & \textbf{42.80} & \textbf{29.43} & \textbf{44.38} & \textbf{36.89} & \textbf{50.69} \\
LaGa~\cite{cen2025laga}
  & 32.20 & 47.35 & 37.19 & 53.88 & 43.06 & 61.31 \\
\rowcolor{oursrow}
\quad + VODA
  & \textbf{33.20} & \textbf{49.63} & \textbf{37.21} & \textbf{55.59} & \textbf{45.53} & \textbf{65.24} \\
\bottomrule
\end{tabular}
\endgroup
\par\medskip
\includegraphics[trim={0pt 4pt 0pt 0pt},clip,width=\linewidth]{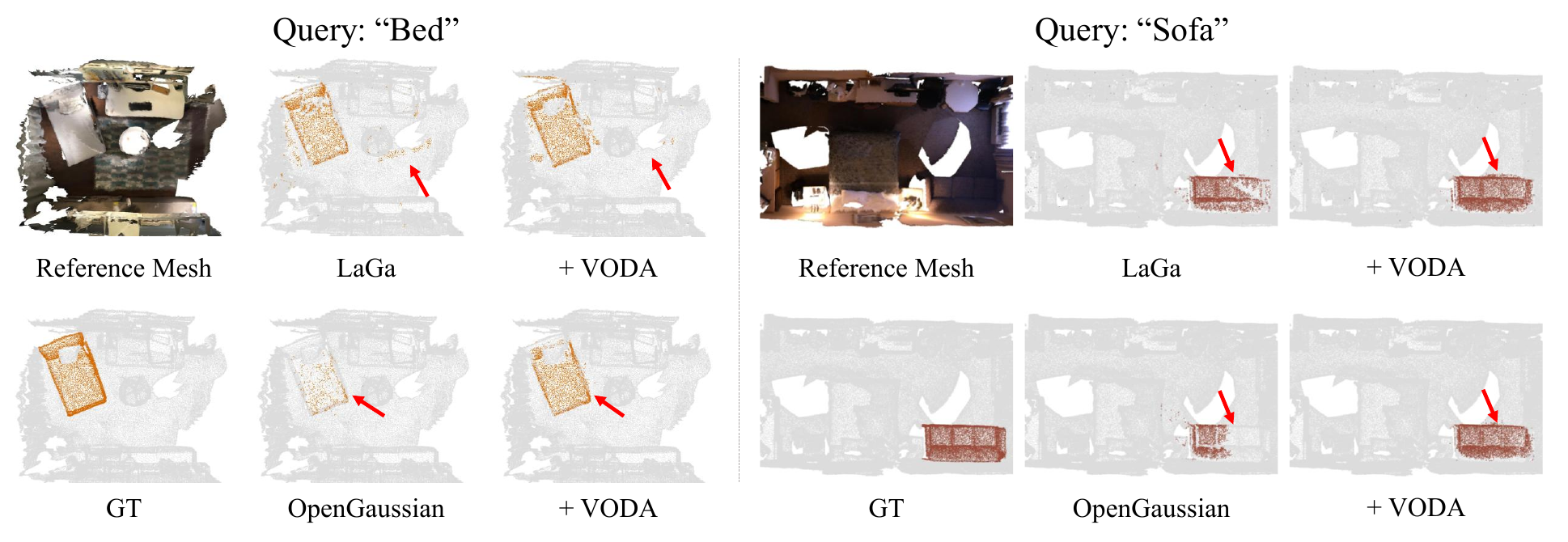}
\captionof{figure}{
Qualitative 3D semantic segmentation on ScanNet.
VODA recovers missing object regions and suppresses spurious semantic assignments for both ``Bed'' and ``Sofa.'' Red arrows highlight representative regions with incomplete or noisy baseline predictions corrected by VODA.
}
\label{fig:scannet_qual}
\end{table}

\begin{table}[t]
\centering
\caption{Volumetric segmentation on Removal-3D.
Rendered-mask mIoU measures spatial overlap, while mRDE measures the depth discrepancy after removing the selected Gaussians.}
\label{table:rde}
\begingroup
\centering
\footnotesize
\setlength{\tabcolsep}{5pt}
\renewcommand{\arraystretch}{1.10}
\begin{tabular}{lcccccccc}
\toprule
\multirow{2}{*}{Methods}
& \multicolumn{2}{c}{Table}
& \multicolumn{2}{c}{Kitchen}
& \multicolumn{2}{c}{Room}
& \multicolumn{2}{c}{\textbf{Mean}} \\
\cmidrule(lr){2-3} \cmidrule(lr){4-5} \cmidrule(lr){6-7} \cmidrule(lr){8-9}
& mIoU$\uparrow$ & mRDE$\downarrow$
& mIoU$\uparrow$ & mRDE$\downarrow$
& mIoU$\uparrow$ & mRDE$\downarrow$
& mIoU$\uparrow$ & mRDE$\downarrow$ \\
\midrule
OpenGaussian~\cite{wu2024opengaussian}
& 57.87 & 43.31
& 37.46 & 29.05
& 63.26 & 45.56
& 52.87 & 39.31 \\
\rowcolor{oursrow} \quad + VODA
& \textbf{65.76} & \textbf{29.39}
& \textbf{52.38} & \textbf{13.00}
& \textbf{64.55} & \textbf{39.64}
& \textbf{60.90} & \textbf{27.34} \\
LaGa~\cite{cen2025laga}
& 77.02 & 43.95
& 66.23 & 22.99
& 84.55 & 43.85
& 75.93 & 36.93 \\
\rowcolor{oursrow} \quad + VODA
& \textbf{79.79} & \textbf{32.25}
& \textbf{70.87} & \textbf{8.16}
& \textbf{86.54} & \textbf{24.56}
& \textbf{79.07} & \textbf{21.66} \\
\midrule
Gaussian Grouping~\cite{ye2024gaussiangrouping}
& 25.57 & 52.65
& 18.52 & 33.92
& 34.59 & 63.57
& 26.23 & 50.05 \\
\rowcolor{oursrow} \quad + VODA
& \textbf{35.94} & \textbf{41.59}
& \textbf{18.66} & \textbf{20.29}
& \textbf{35.93} & \textbf{52.47}
& \textbf{30.18} & \textbf{38.12} \\
ILGS~\cite{jsmbankILGS}
& 27.14 & 50.76
& 26.94 & 32.86
& 40.88 & 66.31
& 31.66 & 49.98 \\
\rowcolor{oursrow} \quad + VODA
& \textbf{37.24} & \textbf{45.19}
& \textbf{31.46} & \textbf{18.40}
& \textbf{47.54} & \textbf{43.34}
& \textbf{38.75} & \textbf{35.65} \\
\bottomrule
\end{tabular}
\endgroup
\par\medskip
\includegraphics[trim={0pt 2pt 0pt 0pt},clip,width=\linewidth]{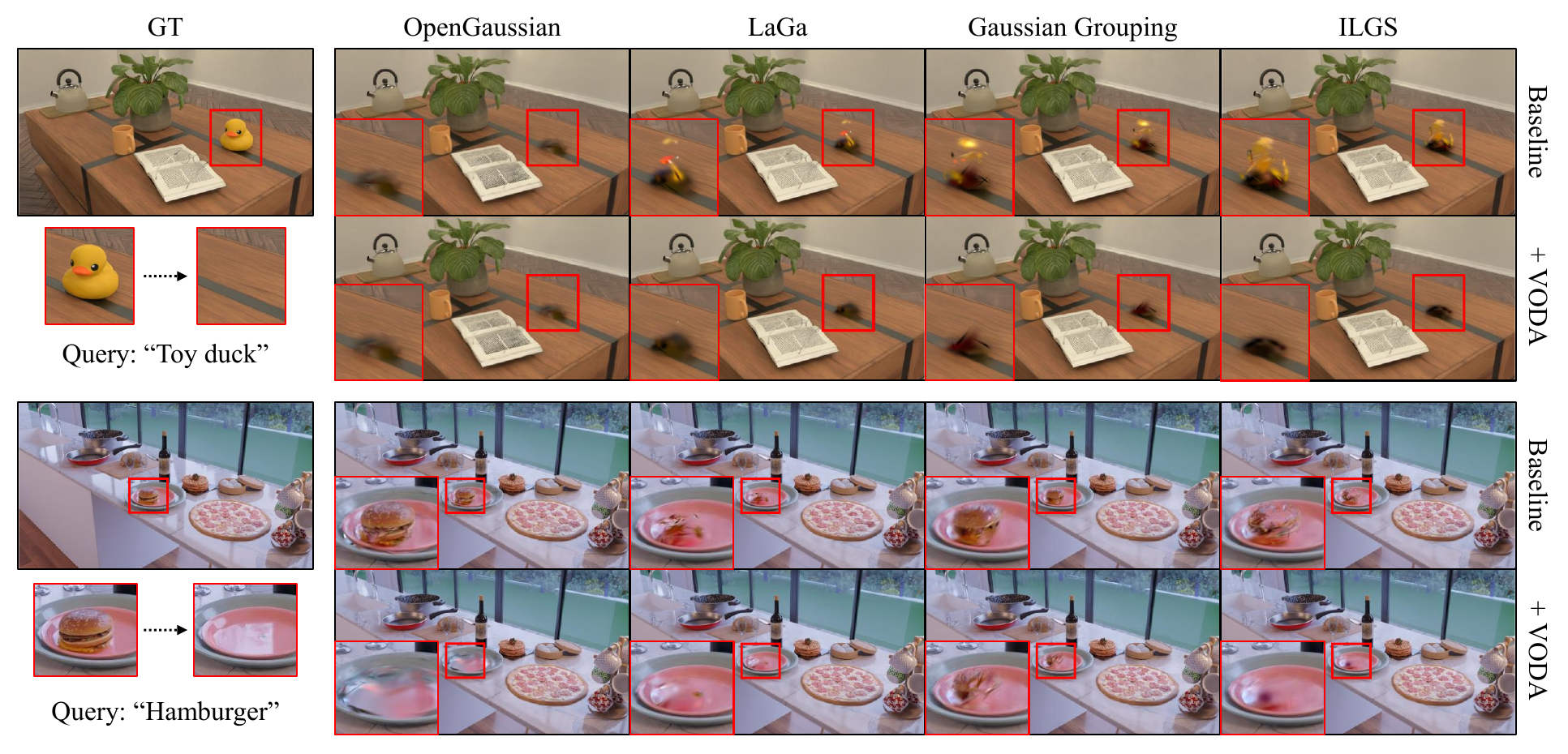}
\captionof{figure}{Qualitative removal on Removal-3D.
Each scene is rendered after removing the queried Gaussians.
VODA leaves less residual target content and better reveals the surface behind the object.}
\label{fig:custom_qual}
\end{table}

\subsection{Comparison on Standard Benchmarks}
\label{sec:exp_compat}

We evaluate VODA on two complementary standard settings probing different aspects of 3D scene understanding: open-vocabulary 3D object selection on LERF and semantic segmentation on ScanNet.

\paragraph{3D Object Selection on LERF.}
Following OpenGaussian, we perform open-vocabulary 3D object selection by selecting a set of Gaussians for each text query and rendering only that set into multiple views.
We compare these rendered masks with the ground-truth object masks from LangSplat~\cite{qin2024langsplat}, reporting mIoU and the percentage of queries whose IoU exceeds $0.25$ (mAcc@0.25).
Because selection precedes rendering, this protocol evaluates a single 3D Gaussian selection consistently across views rather than independently segmenting each rendered image.
Gaussian Grouping and ILGS originally produce 2D segmentations by rendering first and then grouping the rendered features.
We instead evaluate them with the same select-then-render protocol as OpenGaussian and LaGa, so their results differ from those reported in the original papers.
Table~\ref{table:lerf} shows that VODA improves the average mIoU and mAcc@0.25 across all four baselines.
The gains are particularly large for Gaussian Grouping and ILGS, although a few individual scenes show slight regressions.
Figure~\ref{fig:lerf_text_query} shows that VODA recovers missing object parts while reducing spurious selections.

\paragraph{3D Semantic Segmentation on ScanNet.}
Following OpenGaussian, we use 19, 15, and 10 ScanNetv2~\cite{dai2017scannet} categories as text queries and assign each Gaussian to its most similar query.
We transfer these predictions to the ScanNet point cloud and report mIoU and mAcc against the ground-truth semantic labels.
ScanNet complements LERF by evaluating category-level semantic assignment over the entire scene rather than object-level selection from individual text queries.
Table~\ref{table:scannet} shows consistent improvements for both OpenGaussian and LaGa across all three category splits as the semantic label set changes from 19 to 10 classes.
The gains remain consistent across different category splits, suggesting that VODA is not tied to a particular semantic class partition.
Gaussian Grouping and ILGS are not applicable because their features are tied to per-view SAM mask identities, which the ScanNet text-query protocol does not provide.
Figure~\ref{fig:scannet_qual} shows that VODA produces cleaner semantic assignments around object boundaries, with less label leakage across neighboring regions.
However, ScanNet ground-truth labels cover only visible surfaces in the reconstructed point cloud and therefore cannot directly assess occluded object-associated Gaussians, motivating our volumetric benchmark.

\subsection{Comparison on Volumetric Benchmark}
\label{sec:exp_removal}

\paragraph{Removal-3D.}
Standard benchmarks evaluate rendered masks or visible surface labels, but cannot determine whether the selected set covers all Gaussians associated with the 3D object.
In particular, they do not test whether removing the selected Gaussians reveals the surfaces expected after removing the object itself.
We construct Removal-3D, a synthetic benchmark with three indoor scenes (Table, Kitchen, and Room) built in Blender using BlenderKit assets.
For each target object, the benchmark provides paired depth maps of the full scene and the scene after ground-truth object removal.

\paragraph{Removal Depth Error.}
We introduce the Removal Depth Error (RDE) to compare this reference depth change with the change induced by removing the predicted Gaussian set:
\begin{equation}
    \mathrm{RDE}
    =
    \frac{1}{|\Omega|}
    \sum_{p\in\Omega}
    \rho\!\left(\Delta D(p),\,\Delta D_{\mathrm{gt}}(p)\right),
    \label{eq:rde}
\end{equation}
\enlargethispage{1\baselineskip}
\noindent where $\Omega$ denotes the ground-truth target mask region, and $\Delta D(p)$ and $\Delta D_{\mathrm{gt}}(p)$ are the predicted and reference depth changes, respectively.
The function $\rho$ clips the predicted change to the range bounded by the magnitude of the reference change before computing absolute error.
We obtain per-scene mRDE by averaging over viewpoints and target objects, and report the mean across the three scenes as the overall score.
Because RDE is evaluated only within the target object mask, selections outside this region do not affect the score.
We therefore report rendered-mask mIoU alongside RDE as a complementary measure.
RDE measures whether the selected Gaussians remove the target volume as expected, while mIoU measures whether the selection remains spatially aligned with the target in rendered views.
The complete computation is provided in the supplementary.

\paragraph{Volumetric Evaluation on Removal-3D.}
Table~\ref{table:rde} shows consistent reductions in mRDE for every baseline and scene, together with improved mIoU.
The improvement holds across both feature-learning paradigms, demonstrating improved Gaussian-level selection completeness on Removal-3D, a property standard benchmarks leave unmeasured.
The mIoU gains suggest that lower mRDE is not achieved by simply expanding the selected set.
Figure~\ref{fig:custom_qual} complements the LERF and ScanNet visualizations by rendering each scene after removing the queried Gaussians.
For the ``toy duck'' and ``hamburger'' queries, the baselines leave residual target content, indicating that some object Gaussians remain unselected.
VODA leaves less residual content and better reveals the occluded surface.

\subsection{Ablation Study}
\label{sec:exp_ablation}

We ablate supervision delivery, the pixel-level push, and target aggregation using OpenGaussian and Gaussian Grouping as representatives of self-supervised and supervised feature learning.

\begin{table}[htp]
\centering
\footnotesize
\renewcommand{\arraystretch}{1.15}
\begin{minipage}[t]{0.62\textwidth}
\centering
\captionof{table}{Ablation on supervision type and pixel-level push.}
\label{table:supervision_push_ablation}
\setlength{\tabcolsep}{4.0pt}
\resizebox{\linewidth}{!}{
\begin{tabular}{ccc|cc|cc}
\toprule
\multicolumn{2}{c}{Supervision} & \multirow{2}{*}{Pixel-Push}
& \multicolumn{2}{c|}{LERF} & \multicolumn{2}{c}{Removal-3D} \\
Rendering & Direct &
& mIoU$\uparrow$ & mAcc$\uparrow$ & mIoU$\uparrow$ & mRDE$\downarrow$ \\
\midrule
\multicolumn{7}{l}{\textit{\textcolor{sectionlabel}{\small OpenGaussian}}} \\
\addlinespace[3pt]
\checkmark &            & \checkmark
& 39.06 & 56.18 & 52.15 & 32.37 \\
           & \checkmark &
& 42.57 & \textbf{58.14} & 57.90 & 32.63 \\
           & \checkmark & \checkmark
& \textbf{42.98} & 57.84 & \textbf{60.90} & \textbf{27.34} \\
\midrule
\multicolumn{7}{l}{\textit{\textcolor{sectionlabel}{\small Gaussian Grouping}}} \\
\addlinespace[3pt]
\checkmark &            & \checkmark
& 12.40 & 13.28 & 16.84 & 55.60 \\
           & \checkmark &
& 42.89 & \textbf{76.27} & 29.57 & 38.25 \\
           & \checkmark & \checkmark
& \textbf{43.73} & \textbf{76.27} & \textbf{30.18} & \textbf{38.12} \\
\bottomrule
\end{tabular}
}
\end{minipage}
\hfill
\begin{minipage}[t]{0.34\textwidth}
\centering
\captionof{table}{Target aggregation.}
\label{table:aggregation_ablation}
\setlength{\tabcolsep}{4.2pt}
\resizebox{\linewidth}{!}{
\begin{tabular}{cc|cc}
\toprule
\multicolumn{2}{c|}{Aggregation}
& \multicolumn{2}{c}{LERF} \\
Simple & Alpha & mIoU$\uparrow$ & mAcc$\uparrow$ \\
\midrule
\multicolumn{4}{l}{\textit{\textcolor{sectionlabel}{\small OpenGaussian}}} \\
\addlinespace[10pt]
\checkmark &            & 40.63 & 55.89 \\
           & \checkmark & \textbf{42.98} & \textbf{57.84} \\
\midrule
\multicolumn{4}{l}{\textit{\textcolor{sectionlabel}{\small Gaussian Grouping}}} \\
\addlinespace[10pt]
\checkmark &            & 43.22 & \textbf{76.27} \\
           & \checkmark & \textbf{43.73} & \textbf{76.27} \\
\bottomrule
\end{tabular}
}
\end{minipage}
\end{table}

\paragraph{Supervision delivery.}
We compare VODA's direct loss with a rendering variant that uses the same selected depth clusters and target features, but delivers supervision through feature rendering.
Thus, the two variants differ only in how supervision reaches the selected Gaussians.
Direct supervision consistently performs better across both baselines on LERF and Removal-3D, with the largest difference on Gaussian Grouping, where LERF mIoU rises from 12.40 to 43.73.
This shows that identifying the appropriate depth cluster is not sufficient.
Supervision must also bypass the contribution-weighted rendering path and reach the selected Gaussian features directly.

\paragraph{Pixel-level push.}
We next isolate the pixel-level push loss while retaining direct supervision.
The push loss has little effect on standard LERF performance, but consistently improves volumetric selection on Removal-3D.
This supports its complementary role to direct supervision.
Direct supervision improves Gaussian coverage, while the push loss reduces feature mixing between nearby instances and improves volumetric selection.
Boundary visualizations in the supplementary illustrate this effect.

\paragraph{Target aggregation.}
Finally, we compare uniform averaging with the $\alpha$-weighted aggregation in Eq.~\ref{eq:target_aggregation}.
Alpha weighting yields modest but consistent improvements across both baselines, supporting the choice to place greater trust in pixels where the selected Gaussian has a higher per-pixel alpha.
Overall, the ablation shows that direct supervision provides the primary benefit, while the pixel-level push improves instance separation and $\alpha$ weighting improves target reliability.

\FloatBarrier

\subsection{Analysis}
\label{sec:exp_analysis}

We next test whether the Gaussians additionally recovered by VODA are precisely those that receive weak supervision through the rendering path, and whether simpler reweighting or spatial propagation can address the same failure mode.

\begin{figure}[t]
\centering
\includegraphics[width=0.94\linewidth]{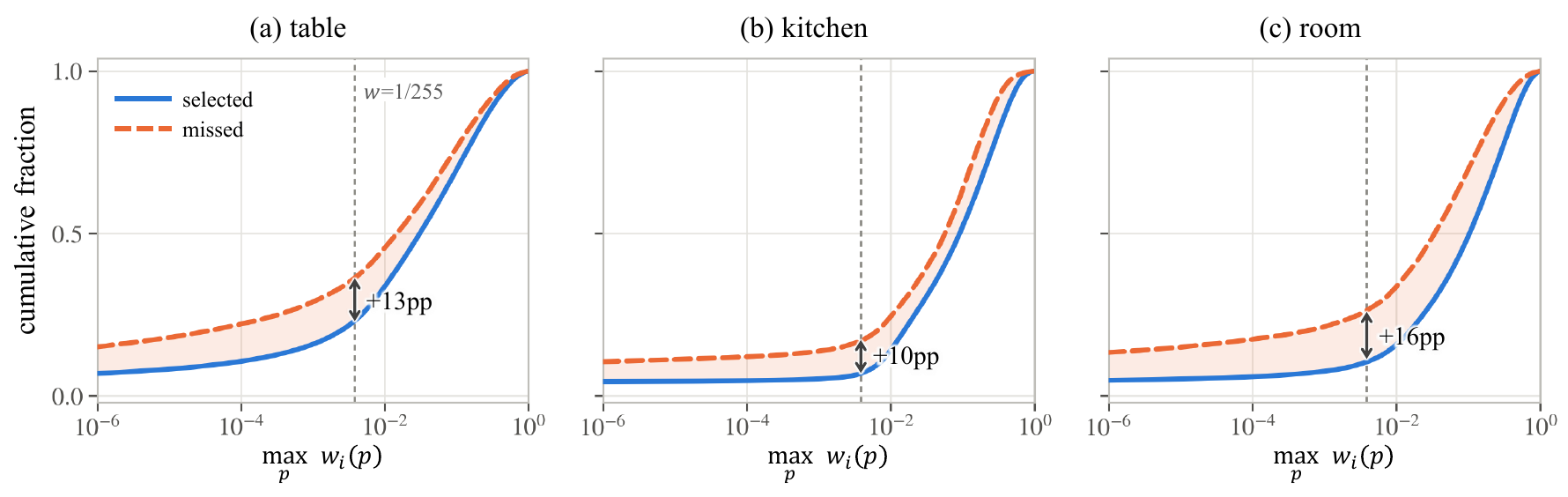}
\caption{
Rendering-contribution CDFs for LaGa on Removal-3D.
VODA-recovered Gaussians have lower contributions than baseline-selected ones; $1/255$ marks the reference threshold.
}
\label{fig:contribution_distribution}
\end{figure}

\begin{table}[t]
\centering
\caption{
Contribution-based alternatives.
Reweighting and propagation yield inconsistent gains, while VODA improves all three metrics for both baselines through direct depth-cluster supervision.
}
\label{table:alternative_baselines}

\footnotesize
\setlength{\tabcolsep}{5pt}
\renewcommand{\arraystretch}{1.10}
\begin{tabular}{lrrr@{\hspace{1.2em}}lrrr}
\toprule
\multicolumn{4}{c}{\textit{\textcolor{sectionlabel}{OpenGaussian}}}
&
\multicolumn{4}{c}{\textit{\textcolor{sectionlabel}{Gaussian Grouping}}} \\
\cmidrule(lr){1-4} \cmidrule(lr){5-8}
\multirow{2}{*}{Methods}
& \multicolumn{1}{c}{LERF}
& \multicolumn{2}{c}{Removal-3D}
& \multirow{2}{*}{Methods}
& \multicolumn{1}{c}{LERF}
& \multicolumn{2}{c}{Removal-3D} \\
\cmidrule(lr){2-2} \cmidrule(lr){3-4}
\cmidrule(lr){6-6} \cmidrule(lr){7-8}
& \multicolumn{1}{c}{mIoU $\uparrow$}
& \multicolumn{1}{c}{mIoU $\uparrow$}
& \multicolumn{1}{c}{mRDE $\downarrow$}
&
& \multicolumn{1}{c}{mIoU $\uparrow$}
& \multicolumn{1}{c}{mIoU $\uparrow$}
& \multicolumn{1}{c}{mRDE $\downarrow$} \\
\midrule
Baseline & 37.12 & 52.87 & 39.3 & Baseline & 22.75 & 26.23 & 50.0 \\
Normalize & 24.61 & 17.1 & 72.7 & Normalize & 34.03 & 1.6 & 48.7 \\
Clamp & 35.94 & 13.4 & 69.4 & Clamp & 31.42 & 1.5 & 52.2 \\
KNN & 30.28 & 17.9 & 73.3 & KNN & 12.24 & 1.5 & 54.7 \\
VODA
& \textbf{42.98} & \textbf{60.90} & \textbf{27.3}
& VODA
& \textbf{43.73} & \textbf{30.18} & \textbf{38.1} \\
\bottomrule
\end{tabular}
\end{table}

\paragraph{Rendering contribution of recovered Gaussians.}
Using LaGa, we compare Gaussians retrieved by the baseline with those missed by the baseline but recovered after adding VODA.
For each Gaussian, we record its maximum rendering contribution $w_i(p)$ over all training pixels and views.
Because the baseline and VODA checkpoints share identical geometry and opacity, as detailed in the supplementary, these contribution statistics are measured under the same scene geometry.
Figure~\ref{fig:contribution_distribution} shows that the recovered Gaussians are consistently shifted toward lower rendering contribution on Removal-3D.
At the $1/255$ cutoff, the fraction of low-contribution Gaussians is 10.2--15.9 percentage points higher in the recovered set across the three scenes.
The supplementary extends this analysis to six scenes, with accumulated-contribution statistics showing the same trend.
This directly supports our diagnosis that VODA preferentially recovers Gaussians with weaker baseline supervision.

\paragraph{Contribution reweighting and spatial propagation.}
Table~\ref{table:alternative_baselines} tests whether the imbalance can be addressed without explicitly assigning supervision through depth clusters.
\emph{Normalize} and \emph{Clamp} rescale the rendering gradient, while \emph{KNN} propagates features from nearby high-contribution Gaussians.
None of these alternatives matches VODA.
All three substantially degrade volumetric performance for OpenGaussian, while for Gaussian Grouping, Normalize improves LERF performance but severely degrades Removal-3D.
These failures reflect a limitation of contribution-based heuristics.
Rendering contribution alone does not indicate whether a Gaussian is associated with the target object.
Reweighting therefore cannot identify the relevant low-contribution Gaussians or recover those that receive no rendered gradient.
Indeed, across the six analyzed scenes, 8.5--18.7\% of recovered Gaussians receive zero contribution in every training view.
KNN avoids this zero-gradient issue, but spatial proximity likewise does not establish object association and may propagate features across object boundaries.
\section{Conclusion}
\label{sec:conclusion}

VODA closes the gap between rendered-mask accuracy and volumetric completeness by grouping contributors by depth and directly supervising every Gaussian in selected clusters.
We introduce Removal-3D and RDE to measure selection completeness from depth changes after object removal.
Across four baselines, VODA improves applicable standard benchmarks and both mRDE and rendered-mask mIoU on Removal-3D.
Reweighting and spatial propagation fail to match VODA, suggesting that reaching low-contribution Gaussians matters more than scaling rendered gradients.

\paragraph{Limitations.}
\label{sec:limitations}
Depth clustering may merge contacting objects or fragment objects with sparse interiors, and the pixel-level push only partly addresses these cases.
Because occluded Gaussians lack feature-level ground truth, we assess their supervision indirectly through downstream selection and removal.
Exact volumetric evaluation requires paired full-scene and object-removed depth maps, available only for the three synthetic Removal-3D scenes.

\clearpage
\section*{Acknowledgments}
This work was supported by the Artificial Intelligence Graduate School Program, Yonsei University (RS-2020-II201361), Institute of Information \& communications Technology Planning \& Evaluation (IITP) grants funded by the Korea government (MSIT) (RS-2024-00457882) and the Global AI Frontier Lab International Collaborative Research (RS-2024-00469482 \& RS-2024-00509279).

\medskip
{
\small
\bibliographystyle{plainnat}
\bibliography{main}

@article{kerbl20233dgs,
  title   = {{3D Gaussian Splatting} for Real-Time Radiance Field Rendering},
  author  = {Kerbl, Bernhard and Kopanas, Georgios and Leimk{\"u}hler, Thomas and Drettakis, George},
  journal = {ACM Transactions on Graphics},
  volume  = {42},
  number  = {4},
  year    = {2023},
  publisher = {ACM}
}

@inproceedings{caron2021dino,
  title     = {Emerging Properties in Self-Supervised Vision Transformers},
  author    = {Caron, Mathilde and Touvron, Hugo and Misra, Ishan and J{\'e}gou, Herv{\'e} and Mairal, Julien and Bojanowski, Piotr and Joulin, Armand},
  booktitle = {IEEE/CVF International Conference on Computer Vision (ICCV)},
  year      = {2021}
}

@inproceedings{dai2017scannet,
  title     = {{ScanNet}: Richly-Annotated {3D} Reconstructions of Indoor Scenes},
  author    = {Dai, Angela and Chang, Angel X. and Savva, Manolis and Halber, Maciej and Funkhouser, Thomas and Nie{\ss}ner, Matthias},
  booktitle = {IEEE/CVF Conference on Computer Vision and Pattern Recognition (CVPR)},
  year      = {2017}
}

@inproceedings{kerr2023lerf,
  title     = {{LERF}: Language Embedded Radiance Fields},
  author    = {Kerr, Justin and Kim, Chung Min and Goldberg, Ken and Kanazawa, Angjoo and Tancik, Matthew},
  booktitle = {IEEE/CVF International Conference on Computer Vision (ICCV)},
  year      = {2023}
}

@inproceedings{zhi2021semanticnerf,
  title     = {In-Place Scene Labelling and Understanding with Implicit Scene Representation},
  author    = {Zhi, Shuaifeng and Laidlow, Tristan and Leutenegger, Stefan and Davison, Andrew J.},
  booktitle = {IEEE/CVF International Conference on Computer Vision (ICCV)},
  year      = {2021}
}

@inproceedings{tschernezki2022n3f,
  title     = {{Neural Feature Fusion Fields}: {3D} Distillation of Self-Supervised {2D} Image Representations},
  author    = {Tschernezki, Vadim and Laina, Iro and Larlus, Diane and Vedaldi, Andrea},
  booktitle = {International Conference on {3D} Vision (3DV)},
  year      = {2022}
}

@inproceedings{kobayashi2022dff,
  title     = {Decomposing {NeRF} for Editing via Feature Field Distillation},
  author    = {Kobayashi, Sosuke and Matsumoto, Eiichi and Sitzmann, Vincent},
  booktitle = {Advances in Neural Information Processing Systems (NeurIPS)},
  year      = {2022}
}

@inproceedings{siddiqui2023panopticlifting,
  title     = {Panoptic Lifting for {3D} Scene Understanding with Neural Fields},
  author    = {Siddiqui, Yawar and Porzi, Lorenzo and Bul{\`o}, Samuel Rota and M{\"u}ller, Norman and Nie{\ss}ner, Matthias and Dai, Angela and Kontschieder, Peter},
  booktitle = {IEEE/CVF Conference on Computer Vision and Pattern Recognition (CVPR)},
  year      = {2023}
}

@inproceedings{bhalgat2023contrastivelift,
  title     = {Contrastive Lift: {3D} Object Instance Segmentation by Slow-Fast Contrastive Fusion},
  author    = {Bhalgat, Yash and Laina, Iro and Henriques, Jo{\~a}o F. and Zisserman, Andrew and Vedaldi, Andrea},
  booktitle = {Advances in Neural Information Processing Systems (NeurIPS)},
  year      = {2023}
}

@inproceedings{liu20233dovs,
  title     = {Weakly Supervised {3D} Open-vocabulary Segmentation},
  author    = {Liu, Kunhao and Zhan, Fangneng and Zhang, Jiahui and Xu, Muyu and Yu, Yingchen and El Saddik, Abdulmotaleb and Theobalt, Christian and Xing, Eric and Lu, Shijian},
  booktitle = {Advances in Neural Information Processing Systems (NeurIPS)},
  year      = {2023}
}

@inproceedings{cen2023sa3d,
  title     = {Segment Anything in {3D} with {NeRFs}},
  author    = {Cen, Jiazhong and Zhou, Zanwei and Fang, Jiemin and Yang, Chen and Shen, Wei and Xie, Lingxi and Jiang, Dongsheng and Zhang, Xiaopeng and Tian, Qi},
  booktitle = {Advances in Neural Information Processing Systems (NeurIPS)},
  year      = {2023}
}

@inproceedings{kim2024garfield,
  title     = {{GARField}: Group Anything with Radiance Fields},
  author    = {Kim, Chung Min and Wu, Mingxuan and Kerr, Justin and Goldberg, Ken and Tancik, Matthew and Kanazawa, Angjoo},
  booktitle = {IEEE/CVF Conference on Computer Vision and Pattern Recognition (CVPR)},
  year      = {2024}
}

@inproceedings{ying2024omniseg3d,
  title     = {{OmniSeg3D}: Omniversal {3D} Segmentation via Hierarchical Contrastive Learning},
  author    = {Ying, Haiyang and Yin, Yixuan and Zhang, Jinzhi and Wang, Fan and Yu, Tao and Huang, Ruqi and Fang, Lu},
  booktitle = {IEEE/CVF Conference on Computer Vision and Pattern Recognition (CVPR)},
  year      = {2024}
}

@inproceedings{bhalgat2024n2f2,
  title     = {{N2F2}: Hierarchical Scene Understanding with Nested Neural Feature Fields},
  author    = {Bhalgat, Yash and Laina, Iro and Henriques, Jo{\~a}o F. and Zisserman, Andrew and Vedaldi, Andrea},
  booktitle = {European Conference on Computer Vision (ECCV)},
  year      = {2024}
}

@inproceedings{ye2024gaussiangrouping,
  title     = {Gaussian Grouping: Segment and Edit Anything in {3D} Scenes},
  author    = {Ye, Mingqiao and Danelljan, Martin and Yu, Fisher and Ke, Lei},
  booktitle = {European Conference on Computer Vision (ECCV)},
  year      = {2024}
}

@inproceedings{zhu2025objectgs,
  title     = {{ObjectGS}: Object-aware Scene Reconstruction and Scene Understanding via {G}aussian Splatting},
  author    = {Zhu, Ruijie and Yu, Mulin and Xu, Linning and Jiang, Lihan and Li, Yixuan and Zhang, Tianzhu and Pang, Jiangmiao and Dai, Bo},
  booktitle = {IEEE/CVF International Conference on Computer Vision (ICCV)},
  year      = {2025}
}

@inproceedings{qin2024langsplat,
  title     = {{LangSplat}: {3D} Language {G}aussian Splatting},
  author    = {Qin, Minghan and Li, Wanhua and Zhou, Jiawei and Wang, Haoqian and Pfister, Hanspeter},
  booktitle = {IEEE/CVF Conference on Computer Vision and Pattern Recognition (CVPR)},
  year      = {2024}
}

@inproceedings{shi2024legaussians,
  title     = {Language Embedded {3D} {G}aussians for Open-Vocabulary Scene Understanding},
  author    = {Shi, Jin-Chuan and Wang, Miao and Duan, Hao-Bin and Guan, Shao-Hua},
  booktitle = {IEEE/CVF Conference on Computer Vision and Pattern Recognition (CVPR)},
  year      = {2024}
}

@inproceedings{zhou2024feature3dgs,
  title     = {Feature {3DGS}: Supercharging {3D} {G}aussian Splatting to Enable Distilled Feature Fields},
  author    = {Zhou, Shijie and Chang, Haoran and Jiang, Sicheng and Fan, Zhiwen and Zhu, Zehao and Xu, Dejia and Chari, Pradyumna and You, Suya and Wang, Zhangyang and Kadambi, Achuta},
  booktitle = {IEEE/CVF Conference on Computer Vision and Pattern Recognition (CVPR)},
  year      = {2024}
}

@article{zuo2024fmgs,
  title   = {{FMGS}: Foundation Model Embedded {3D} {G}aussian Splatting for Holistic {3D} Scene Understanding},
  author  = {Zuo, Xingxing and Samangouei, Pouya and Zhou, Yunwen and Di, Yan and Li, Mingyang},
  journal = {International Journal of Computer Vision},
  year    = {2024},
  publisher = {Springer}
}

@inproceedings{qu2024goi,
  title     = {{GOI}: Find {3D} {G}aussians of Interest with an Optimizable Open-vocabulary Semantic-space Hyperplane},
  author    = {Qu, Yansong and Dai, Shaohui and Li, Xinyang and Lin, Jianghang and Cao, Liujuan and Zhang, Shengchuan and Ji, Rongrong},
  booktitle = {ACM International Conference on Multimedia (MM)},
  year      = {2024}
}

@article{li2024langsurf,
  title   = {{LangSurf}: Language-Embedded Surface {G}aussians for {3D} Scene Understanding},
  author  = {Li, Hao and Qin, Roy and Zou, Zhengyu and He, Diqi and Li, Bohan and Dai, Bingquan and Zhang, Dingwen and Han, Junwei},
  journal = {arXiv preprint arXiv:2412.17635},
  year    = {2024}
}

@inproceedings{wu2024opengaussian,
  title     = {{OpenGaussian}: Towards Point-Level {3D} {G}aussian-based Open Vocabulary Understanding},
  author    = {Wu, Yanmin and Meng, Jiarui and Li, Haijie and Wu, Chenming and Shi, Yahao and Cheng, Xinhua and Zhao, Chen and Feng, Haocheng and Ding, Errui and Wang, Jingdong and Zhang, Jian},
  booktitle = {Advances in Neural Information Processing Systems (NeurIPS)},
  year      = {2024}
}

@inproceedings{cen2025laga,
  title     = {Tackling View-Dependent Semantics in {3D} Language {G}aussian Splatting},
  author    = {Cen, Jiazhong and Zhou, Xudong and Fang, Jiemin and Wen, Changsong and Xie, Lingxi and Zhang, Xiaopeng and Shen, Wei and Tian, Qi},
  booktitle = {International Conference on Machine Learning (ICML)},
  year      = {2025}
}

@inproceedings{cen2025saga,
  title     = {Segment Any {3D} {G}aussians},
  author    = {Cen, Jiazhong and Fang, Jiemin and Yang, Chen and Xie, Lingxi and Zhang, Xiaopeng and Shen, Wei and Tian, Qi},
  booktitle = {AAAI Conference on Artificial Intelligence (AAAI)},
  year      = {2025}
}

@inproceedings{choi2024clickgaussian,
  title     = {{Click-Gaussian}: Interactive Segmentation to Any {3D} {G}aussians},
  author    = {Choi, Seokhun and Song, Hyeonseop and Kim, Jaechul and Kim, Taehyeong and Do, Hoseok},
  booktitle = {European Conference on Computer Vision (ECCV)},
  year      = {2024}
}

@inproceedings{lee2025open3drf,
  title     = {Rethinking Open-Vocabulary Segmentation of Radiance Fields in {3D} Space},
  author    = {Lee, Hyunjee and Yun, Youngsik and Bae, Jeongmin and Kim, Seoha and Uh, Youngjung},
  booktitle = {AAAI Conference on Artificial Intelligence (AAAI)},
  year      = {2025}
}

@inproceedings{spinnerf,
      title={{SPIn-NeRF}: Multiview Segmentation and Perceptual Inpainting with Neural Radiance Fields}, 
      author={Ashkan Mirzaei and Tristan Aumentado-Armstrong and Konstantinos G. Derpanis and Jonathan Kelly and Marcus A. Brubaker and Igor Gilitschenski and Alex Levinshtein},
      year={2023},
      booktitle={CVPR},
}

@InProceedings{wu2025aurafusion,
  author    = {Wu, Chung-Ho and Chen, Yang-Jung and Chen, Ying-Huan and Lee, Jie-Ying and Ke, Bo-Hsu and Mu, Chun-Wei Tuan and Huang, Yi-Chuan and Lin, Chin-Yang and Chen, Min-Hung and Lin, Yen-Yu and Liu, Yu-Lun},
  title     = {AuraFusion360: Augmented Unseen Region Alignment for Reference-based 360deg Unbounded Scene Inpainting},
  booktitle = {Proceedings of the Computer Vision and Pattern Recognition Conference (CVPR)},
  month     = {June},
  year      = {2025},
  pages     = {16366-16376}
}

@inproceedings{chacko2025lbg,
  title     = {Lifting by {G}aussians: A Simple, Fast and Flexible Method for {3D} Instance Segmentation},
  author    = {Chacko, Rohan and H{\"a}ni, Nicolai and Khaliullin, Eldar and Sun, Lin and Lee, Douglas},
  booktitle = {IEEE/CVF Winter Conference on Applications of Computer Vision (WACV)},
  year      = {2025}
}

@inproceedings{kim2025drsplat,
  title     = {{Dr.~Splat}: Directly Referring {3D} {G}aussian Splatting via Direct Language Embedding Registration},
  author    = {Kim, Jun-Seong and Kim, GeonU and Kim, Yu-Ji and Wang, Yu-Chiang Frank and Choe, Jaesung and Oh, Tae-Hyun},
  booktitle = {IEEE/CVF Conference on Computer Vision and Pattern Recognition (CVPR)},
  year      = {2025}
}

@article{jeong2025qrender,
  title   = {Quantile Rendering: Efficiently Embedding High-dimensional Feature on {3D} {G}aussian Splatting},
  author  = {Jeong, Yoonwoo and Sun, Cheng and Wang, Frank and Cho, Minsu and Choe, Jaesung},
  journal = {arXiv preprint arXiv:2512.20927},
  year    = {2025}
}

@inproceedings{jsmbankILGS,
  title={Identity-aware Language Gaussian Splatting for Open-vocabulary 3D Semantic Segmentation},
  author={SungMin Jang and Wonjun Kim},
  booktitle={Proceedings of the International Conference on Computer Vision},
  year={2025}
}

@inproceedings{lu2025segment,
  title={Segment then Splat: Unified 3D Open-Vocabulary Segmentation via Gaussian Splatting},
  author={Lu, Yiren and Zhou, Yunlai and Qiao, Yiran and Song, Chaoda and Liang, Tuo and Ma, Jing and Wang, Huan and Yin, Yu},
  booktitle = {Advances in Neural Information Processing Systems (NeurIPS)},
  year      = {2025}
}
}


\clearpage
\appendix

\section{Appendix}

%
%

\subsection{Implementation Details}
\label{sec:imple_detail}

We implement VODA in PyTorch on top of the official codebase of each baseline and run the main experiments on a single NVIDIA RTX A6000 GPU.
VODA changes only feature learning: data preparation, scene reconstruction, the feature representation, and the inference-time Gaussian selection procedure remain unchanged.
At an interval of $T_{\mathrm{int}}$ iterations, we run the depth-aware rasterizer and add $\mathcal{L}_{\mathrm{direct}}$ and $\mathcal{L}_{\mathrm{push}}$ to the native baseline objective.

\paragraph{Baseline integration.}
For OpenGaussian~\cite{wu2024opengaussian} and LaGa~\cite{cen2025laga}, we enable VODA when instance-feature learning begins, after scene reconstruction, with $T_{\mathrm{int}}=5$.
For OpenGaussian, we map the Gaussian features and rendered targets to $[0,1]$ before applying the squared-$\ell_2$ direct loss.
Subsequent codebook and language-association stages remain unchanged.

LaGa learns a 32-dimensional feature with a cosine-based contrastive objective.
We therefore replace the squared-$\ell_2$ distance in Eq.~\ref{eq:dir_loss} with cosine distance.
Here, $\hat{\mathbf{F}}_i$ is the $\alpha_i(p)$-weighted aggregate of pixel-level rendered feature targets for $g_i$, as defined in Eq.~\ref{eq:target_aggregation}:
\begin{equation}
    \mathcal{L}_{\mathrm{direct}}^{\mathrm{LaGa}}
    \;=\;
    \frac{1}{|\mathcal{G}_{\mathrm{sel}}|}
    \sum_{g_i\in\mathcal{G}_{\mathrm{sel}}}
    \Bigl(
      1
      \;-\;
      \frac{\hat{\mathbf{F}}_i^{\!\top}\,\mathbf{f}_i}
           {\lVert\hat{\mathbf{F}}_i\rVert_2\,\lVert\mathbf{f}_i\rVert_2 + \epsilon}
    \Bigr).
    \label{eq:dir_laga}
\end{equation}

Gaussian Grouping~\cite{ye2024gaussiangrouping} and ILGS~\cite{jsmbankILGS} learn identity features jointly with scene reconstruction.
For the reported experiments, we retain their standard training and append a $5{,}000$-iteration feature fine-tuning stage with $T_{\mathrm{int}}=1$.
Geometry, appearance, and the identity classifier are frozen, while the existing Gaussian identity features are optimized with the original loss and the two VODA losses.
Gaussian Grouping retains its original 3D consistency term.
For ILGS, the CLIP-aligned semantic feature is also frozen, and VODA is applied only to the identity feature used for instance grouping.

\paragraph{Hyperparameters.}
Across all four baselines, we use $\gamma=15$ and clip the adaptive depth threshold to $[\tau_{\min},\tau_{\max}]=[0.05,1.0]$.
We set $\lambda_{\mathrm{direct}}=0.5$, $\lambda_{\mathrm{push}}=0.1$, and the push margin to $m=0.3$.
These values are shared without per-baseline tuning, while only the integration schedule differs.

\subsection{Removal-3D Benchmark and RDE Computation}
\label{sec:removal3d_detail}

\paragraph{Dataset.}
Removal-3D contains three synthetic indoor scenes---Table, Kitchen, and Room---constructed in Blender using BlenderKit assets.
For every target, the benchmark provides the full scene and a counterpart with that object removed, along with object masks, depth maps, and camera parameters estimated with COLMAP.

\paragraph{Depth rendering and alignment.}
For target object $o$ in view $f$, let $D^{\mathrm{full}}_{\mathrm{gt}}(u)$ be the benchmark depth at pixel $u$ in the full scene, and let $D^{\mathrm{rem},(o)}_{\mathrm{gt}}(u)$ be the depth in the paired scene with $o$ removed.
Their signed difference is
\begin{equation}
    \Delta D_{\mathrm{gt}}^{(o)}(u)
    \;=\;
    D^{\mathrm{full}}_{\mathrm{gt}}(u) - D^{\mathrm{rem},(o)}_{\mathrm{gt}}(u).
    \label{eq:gt_delta}
\end{equation}
We render predicted depth before and after removing the selected Gaussian set using the standard alpha-compositing weights:
\begin{equation}
    D_{\mathrm{pred}}(u)
    \;=\;
    \sum_{i\in\mathcal{G}(u)} T_i(u)\,\alpha_i(u)\,z_i(u),
    \label{eq:depth_render}
\end{equation}
where $z_i(u)$ is the camera-space depth of $g_i$.
The predicted depth is not calibrated to the benchmark depth, so we fit a global scale and offset using the full-scene prediction and its ground-truth depth:
\begin{equation}
    (s^{*},\,t^{*})
    \;=\;
    \arg\min_{s,t}\;
    \sum_{u\in\mathcal{V}}\!
    \bigl(s\,D^{\mathrm{full}}_{\mathrm{pred}}(u)+t-D^{\mathrm{full}}_{\mathrm{gt}}(u)\bigr)^{2},
    \label{eq:scaleshift}
\end{equation}
where $\mathcal{V}=\{u: D^{\mathrm{full}}_{\mathrm{pred}}(u)>0 \,\land\, D^{\mathrm{full}}_{\mathrm{gt}}(u)>0\}$.
We apply the same transform to the predicted removal depth.
Subtracting the aligned depths cancels the offset, giving $\Delta D_{\mathrm{pred}}^{(o)}(u)=s^*(D^{\mathrm{full}}_{\mathrm{pred}}(u)-D^{\mathrm{rem},(o)}_{\mathrm{pred}}(u))$.

\paragraph{Removal Depth Error.}
We compute the RDE over the ground-truth mask $m_o$ of target object $o$, using the pixels $\Omega_o=\{u:m_o(u)=1\}$.
Before computing the mean absolute error, we clip the signed predicted depth change to the symmetric interval defined by the reference magnitude:
\begin{equation}
    \widetilde{\Delta D}_{\mathrm{pred}}^{(o)}(u)
    =\mathrm{clip}\!\left(
    \Delta D_{\mathrm{pred}}^{(o)}(u),
    -|\Delta D_{\mathrm{gt}}^{(o)}(u)|,
    |\Delta D_{\mathrm{gt}}^{(o)}(u)|\right).
    \label{eq:clip}
\end{equation}
The reported RDE is
\begin{equation}
    \mathrm{RDE}(o,f)=
    \frac{1}{|\Omega_o|}\sum_{u\in\Omega_o}
    \left|\widetilde{\Delta D}_{\mathrm{pred}}^{(o)}(u)
    -\Delta D_{\mathrm{gt}}^{(o)}(u)\right|.
    \label{eq:rde_refine}
\end{equation}

\paragraph{Evaluation and aggregation.}
For each scene, we evaluate the eight predefined test views (\texttt{frame\_00}, \texttt{frame\_20}, $\ldots$, \texttt{frame\_140}).
We average the per-view RDE over views for each object, macro-average over objects to obtain each scene's mRDE, and report the mean across scenes:
\begin{align}
    \text{per-object RDE}
    &= \tfrac{1}{|\mathcal{F}_o|}\sum_{f\in\mathcal{F}_o}\mathrm{RDE}(o, f), \\
    \text{per-scene mRDE}
    &= \tfrac{1}{|\mathcal{O}_s|}\sum_{o\in\mathcal{O}_s}\text{per-object RDE}(o), \\
    \text{total mRDE}
    &= \tfrac{1}{|\mathcal{S}|}\sum_{s\in\mathcal{S}}\text{per-scene mRDE}(s).
\end{align}
We also report rendered-mask mIoU on the same views by projecting $\hat{\mathcal{G}}_o$ and comparing it with the ground-truth mask $m_o$.

\FloatBarrier
\subsection{Additional Baseline Comparison and Positioning}
\label{sec:additional_recent_works}

\paragraph{Reconstruction-time correspondence: Segment then Splat.}
Segment then Splat (STS)~\cite{lu2025segment} and VODA both seek object-consistent Gaussian representations, but introduce object identity at different stages.
STS partitions Gaussians into object-specific sets using pre-obtained masks and optimizes the scene under this fixed correspondence.
VODA instead operates on an already-reconstructed scene: it adds direct object-level supervision to the baseline's features while leaving geometry, opacity, and appearance unchanged.
STS therefore establishes object-specific Gaussian sets during reconstruction, whereas VODA improves feature-based object assignment after reconstruction.

\paragraph{Volumetric segmentation and reconstruction quality.}
We evaluate the public STS implementation under the same Removal-3D protocol (Table~\ref{table:sts_removal}).
STS achieves higher selection mIoU than Gaussian Grouping (38.37 vs.\ 26.23), but has a higher mRDE than either unmodified baseline (62.10 vs.\ 39.31 and 50.05).
In contrast, VODA improves both metrics for each baseline without retraining its geometry.
Table~\ref{table:sts_reconstruction} also reports full-scene reconstruction quality averaged over the three scenes.
VODA leaves these metrics unchanged because it modifies only object features.
In this setting, STS has lower PSNR and SSIM and higher LPIPS than the two baselines.
These results show that reconstruction-time correspondence alone does not ensure volumetrically complete object selection.
However, STS and the feature-learning baselines use different reconstructed scenes, so this comparison does not isolate object assignment from reconstruction quality.
The baseline--VODA comparisons keep the reconstructed scene fixed and isolate the effect of feature supervision.

\paragraph{Complementary scope and trade-offs.}
Fixed mask-based correspondence can propagate errors from partial views or ambiguous boundaries into both object assignment and geometry optimization.
This dependence makes reliable masks and cross-view tracking important for reconstruction-time methods.
VODA decouples feature supervision from reconstruction and can be applied to pre-trained scenes, but cannot repair errors in the geometry it inherits.
STS is attractive when a reliable initial object decomposition is available, while VODA addresses incomplete selection in existing rendering-based feature-learning pipelines.
We therefore view the approaches as complementary rather than claiming that either paradigm universally dominates the other.

\begin{table}[htbp]
\centering
\begin{minipage}[t]{0.49\linewidth}
\centering
\small
\setlength{\tabcolsep}{4pt}
\renewcommand{\arraystretch}{1.08}
\caption{Object selection on Removal-3D.}
\label{table:sts_removal}
\begin{tabular}{@{}lcc@{}}
\toprule
Method & mIoU $\uparrow$ & mRDE $\downarrow$ \\
\midrule
OpenGaussian & 52.87 & 39.31 \\
OpenGaussian + VODA & \textbf{60.90} & \textbf{27.34} \\
Gaussian Grouping & 26.23 & 50.05 \\
Gaussian Grouping + VODA & 30.18 & 38.12 \\
Segment then Splat & 38.37 & 62.10 \\
\bottomrule
\end{tabular}
\end{minipage}\hfill
\begin{minipage}[t]{0.49\linewidth}
\centering
\small
\setlength{\tabcolsep}{4pt}
\renewcommand{\arraystretch}{1.08}
\caption{RGB reconstruction on Removal-3D.}
\label{table:sts_reconstruction}
\begin{tabular}{@{}lccc@{}}
\toprule
Method & PSNR $\uparrow$ & SSIM $\uparrow$ & LPIPS $\downarrow$ \\
\midrule
OpenGaussian & 33.4 & 0.925 & 0.105 \\
Gaussian Grouping & 34.7 & 0.954 & 0.069 \\
Segment then Splat & 28.2 & 0.884 & 0.237 \\
\bottomrule
\end{tabular}
\end{minipage}
\end{table}

\FloatBarrier
\subsection{Diagnostic Analyses}
\label{sec:diagnostic_analyses}

\paragraph{Scene-level contribution statistics.}
The main-paper analysis (Fig.~\ref{fig:contribution_distribution}) shows that Gaussians recovered by VODA have lower maximum rendering contributions on Removal-3D.
We extend this analysis across six scenes from LERF and Removal-3D using accumulated-contribution statistics.

For each Gaussian, we compute its accumulated contribution by summing $w_i(p)=T_i(p)\alpha_i(p)$ over training pixels and views.
We also record its maximum per-pixel contribution.
We compare baseline-retrieved Gaussians (\emph{selected}) with those missed by the baseline but retrieved after adding VODA (\emph{missed}).
Both sets are measured in the same reconstructed scene.
Because the baseline and VODA checkpoints share geometry and opacity, each Gaussian has identical rendering contributions across the two checkpoints.
Across Removal-3D scenes, VODA excludes 1.9--6.8\% of the Gaussians selected by the baseline.
Excluding these Gaussians changes the cutoff fractions by at most 0.3 percentage points.

Table~\ref{table:contribution_statistics} shows that, across all six scenes, the missed set has lower median accumulated contribution and higher fractions of low-maximum and zero-contribution Gaussians than the selected set.

\begin{table}[htbp]
\centering
\footnotesize
\setlength{\tabcolsep}{5pt}
\renewcommand{\arraystretch}{1.08}
\caption{
Rendering-contribution statistics for LaGa.
Selected denotes baseline-retrieved Gaussians, and Missed denotes those recovered only with VODA.
The median is computed over accumulated training-view contributions.
The final columns give fractions below the maximum-contribution cutoff and with zero contribution in every view.
}
\label{table:contribution_statistics}
\begin{tabular}{lllrrr}
\toprule
Dataset & Scene & Set & Median $\sum w$ & $\max w < 1/255$ (\%) & Zero (\%) \\
\midrule
\multirow{6}{*}{LERF}
& \multirow{2}{*}{Figurines} & Selected   & 119.87 & 25.1 & 10.6 \\
&                            & Missed     & 0.23   & 59.3 & 18.7 \\
\cmidrule(lr){2-6}
& \multirow{2}{*}{Ramen}     & Selected   & 29.70 & 15.1 & 6.4 \\
&                            & Missed     & 9.75  & 21.5 & 10.9 \\
\cmidrule(lr){2-6}
& \multirow{2}{*}{Teatime}   & Selected   & 57.66 & 7.5  & 3.7 \\
&                            & Missed     & 15.48 & 18.6 & 8.5 \\
\midrule
\multirow{6}{*}{Removal-3D}
& \multirow{2}{*}{Kitchen}   & Selected   & 20.33 & 6.9  & 4.4 \\
&                            & Missed     & 4.95  & 17.1 & 10.3 \\
\cmidrule(lr){2-6}
& \multirow{2}{*}{Room}      & Selected   & 26.50 & 10.5 & 4.7 \\
&                            & Missed     & 7.30  & 26.4 & 12.8 \\
\cmidrule(lr){2-6}
& \multirow{2}{*}{Table}     & Selected   & 8.71  & 23.3 & 6.6 \\
&                            & Missed     & 1.72  & 36.5 & 14.3 \\
\bottomrule
\end{tabular}
\end{table}

\begin{figure}[tbp]
\centering
\includegraphics[width=0.90\linewidth]{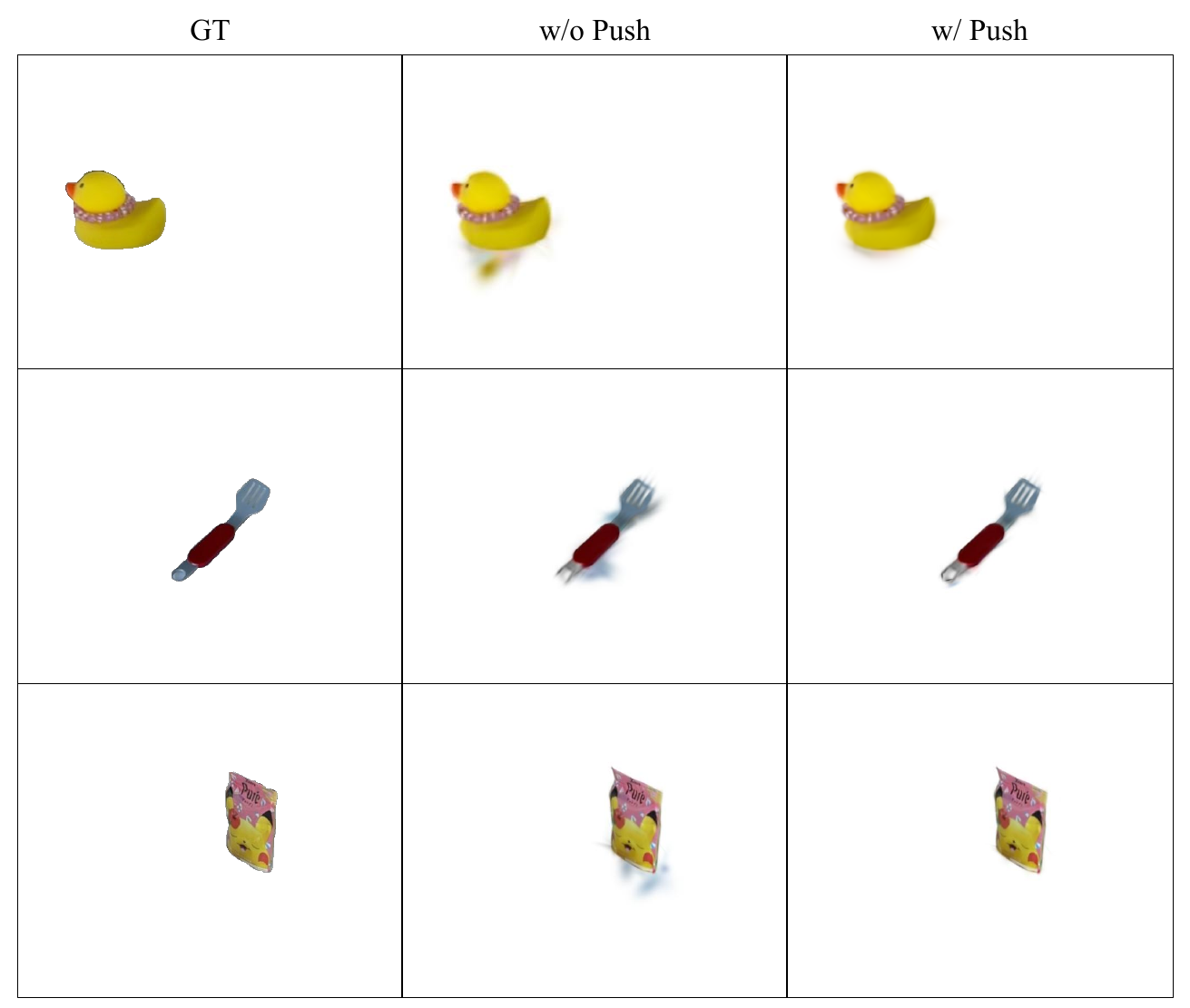}
\caption{
Effect of the pixel-level push loss.
Ground truth (left), VODA without push (middle), and VODA with push (right).
The push loss sharpens adjacent-instance boundaries.
}
\label{fig:pixel-push}
\end{figure}

\paragraph{Pixel-level push.}
The pixel-level push sharpens boundaries between adjacent instances that depth clustering alone may not separate.
Figure~\ref{fig:pixel-push} compares the rendered masks with and without this loss.

\FloatBarrier
\subsection{Robustness and Efficiency}
\label{sec:robustness_efficiency}

\paragraph{Depth-threshold sensitivity.}
We vary $\gamma$ and the threshold clamps using Gaussian Grouping on LERF (Tables~\ref{table:gamma_sensitivity} and~\ref{table:clamp_sensitivity}).
Across the sweeps, mIoU spans 0.43 and 1.03 percentage points, respectively.
The default $\gamma=15$ is near-best.
Raising or removing $\tau_{\max}$ lowers mIoU, while changes to $\tau_{\min}$ have smaller effects in the tested range.
Dashes indicate omitted clamp bounds.

\begin{table}[htbp]
\centering
\begin{minipage}[t]{0.48\linewidth}
\centering
\renewcommand{\arraystretch}{1.08}
\caption{Split sensitivity on LERF.}
\label{table:gamma_sensitivity}
\begin{tabular*}{\linewidth}{@{\extracolsep{\fill}}cc}
\toprule
$\gamma$ & mIoU $\uparrow$ \\
\midrule
5 & 43.40 \\
10 & 43.58 \\
15 (default) & 43.73 \\
20 & 43.83 \\
25 & 43.51 \\
30 & 43.70 \\
\bottomrule
\end{tabular*}
\end{minipage}\hfill
\begin{minipage}[t]{0.48\linewidth}
\centering
\renewcommand{\arraystretch}{1.08}
\caption{Threshold clamps on LERF.}
\label{table:clamp_sensitivity}
\begin{tabular*}{\linewidth}{@{\extracolsep{\fill}}ccc}
\toprule
$\tau_{\min}$ & $\tau_{\max}$ & mIoU $\uparrow$ \\
\midrule
0.05 & 1.0 & 43.73 \\
0.025 & 1.0 & 44.13 \\
0.075 & 1.0 & 43.65 \\
0.05 & 0.5 & 43.97 \\
0.05 & 1.5 & 43.19 \\
-- & 1.0 & 44.13 \\
0.05 & -- & 43.10 \\
-- & -- & 43.17 \\
\bottomrule
\end{tabular*}
\end{minipage}
\end{table}

\paragraph{Injection timing.}
Table~\ref{table:injection_timing} tests whether VODA can be optimized jointly with reconstruction.
Enabling VODA from the start of training improves both mIoU and mRDE over the baseline (33.1 vs. 26.4 and 32.2 vs. 46.4), showing that VODA need not wait for geometry training to finish.
Enabling it at 10K or 20K iterations also outperforms waiting until geometry training ends at 30K on both metrics.
The 10K and 20K settings reach mIoU values of 32.1 and 32.2, compared with 31.1 at 30K, and mRDE values of 24.1 and 25.0, compared with 30.2 at 30K.
The best timing depends on the metric.
Starting at 0 gives the highest mIoU, while starting at 10K gives the lowest mRDE in this sweep.
The main experiments use post-training feature fine-tuning to keep geometry fixed and isolate the effect of feature supervision.
These values come from a separate diagnostic run and differ slightly from the reported results.

\begin{table}[htbp]
\centering
\small
\caption{Sensitivity to the VODA injection point for Gaussian Grouping.}
\label{table:injection_timing}
\begin{tabular}{lrrrrr}
\toprule
 & Baseline & $t=0$ & $t=10$K & $t=20$K & $t=30$K \\
\midrule
mIoU $\uparrow$ & 26.4 & 33.1 & 32.1 & 32.2 & 31.1 \\
mRDE $\downarrow$ & 46.4 & 32.2 & 24.1 & 25.0 & 30.2 \\
\bottomrule
\end{tabular}
\end{table}

\begin{samepage}
\paragraph{Interior versus boundary RDE.}
We split the evaluation mask into an interior and a boundary to test whether VODA improves removal throughout the object or mainly near its silhouette.
We define the interior by eroding the ground-truth mask by five pixels.
It retains 95--97\% of the object mask, and the removed border forms the boundary region.
Table~\ref{table:interior_boundary} shows comparable mRDE reductions in both regions for both baselines, indicating that the gains are not confined to boundary alignment.\par
\end{samepage}

\begin{table}[htbp]
\centering
\small
\caption{Interior and boundary mRDE on Removal-3D.
The interior is obtained by eroding the object mask by five pixels; $\Delta$ denotes the reduction in mRDE relative to the corresponding baseline.}
\label{table:interior_boundary}
\begin{tabular}{lrrrr}
\toprule
Method & Interior $\downarrow$ & Boundary $\downarrow$ & $\Delta$ Interior & $\Delta$ Boundary \\
\midrule
OpenGaussian & 39.22 & 39.09 & -- & -- \\
$+$ VODA ($-$push) & 31.46 & 32.42 & 7.8 & 6.7 \\
$+$ VODA & \textbf{26.81} & \textbf{27.12} & 12.4 & 12.0 \\
\midrule
Gaussian Grouping & 49.80 & 46.82 & -- & -- \\
$+$ VODA ($-$push) & 39.23 & \textbf{34.84} & 10.6 & 12.0 \\
$+$ VODA & \textbf{39.13} & 34.96 & 10.7 & 11.9 \\
\bottomrule
\end{tabular}
\end{table}

\paragraph{Computational overhead.}
Table~\ref{table:runtime} reports timings over 1,000 iterations on LERF-figurines using an RTX A5000, with one baseline from each integration schedule.
Most of the added cost comes from the no-gradient depth-aware rasterizer forward pass.
Target aggregation and the additional backward pass are comparatively small.
The per-iteration overhead applies only while VODA is active.
It adds approximately 10 minutes to the OpenGaussian pipeline and 20 minutes to the Gaussian Grouping pipeline.
VODA adds no inference-time cost because the baseline selection procedure is unchanged.

\begin{table}[htbp]
\centering
\small
\caption{Training overhead on LERF-figurines (RTX A5000).
Per-iteration timings are averaged over 1,000 iterations.
Full-pipeline values give the added training time.}
\label{table:runtime}
\begin{tabular}{lrr}
\toprule
 & OpenGaussian ($T_{\mathrm{int}}=5$) & Gaussian Grouping ($T_{\mathrm{int}}=1$) \\
\midrule
Baseline per iteration & 124.1 ms & 75.7 ms \\
$+$ VODA per iteration & 183.3 ms ($+47.7\%$) & 302.3 ms ($+299\%$) \\
Depth-aware rasterizer & 55.8 ms & 219.9 ms \\
Targets, losses, backward & $\sim$3 ms & $\sim$7 ms \\
Added full-pipeline time & $\sim$10 min & $\sim$20 min \\
\bottomrule
\end{tabular}
\end{table}

\clearpage
\FloatBarrier
\subsection{Run-to-Run Variability}
\label{app:reproducibility}

We repeat VODA training three times on each applicable benchmark while keeping baseline results fixed.
Tables~\ref{table:repeat_lerf}--\ref{table:repeat_removal} report mean $\pm$ sample standard deviation ($n=3$).
The main tables report the first run.

\begin{table}[htbp]
\centering
\small
\setlength{\tabcolsep}{7pt}
\renewcommand{\arraystretch}{1.10}
\caption{Run-to-run variability on LERF (mIoU $\uparrow$).
Mean $\pm\sigma$ is computed over three VODA runs with fixed baseline results.}
\label{table:repeat_lerf}
\begin{tabular}{@{}lrr@{}}
\toprule
Method & Baseline & + VODA (Mean $\pm\sigma$) \\
\midrule
OpenGaussian & 37.12 & $42.29\pm0.80$ \\
LaGa & 56.80 & $60.17\pm0.78$ \\
Gaussian Grouping & 22.75 & $44.07\pm0.31$ \\
ILGS & 30.00 & $49.27\pm0.52$ \\
\bottomrule
\end{tabular}
\end{table}

\begin{table}[htbp]
\centering
\small
\setlength{\tabcolsep}{7pt}
\renewcommand{\arraystretch}{1.10}
\caption{Run-to-run variability on ScanNet (mIoU $\uparrow$).
Mean $\pm\sigma$ is computed over three VODA runs for each class split.}
\label{table:repeat_scannet}
\begin{tabular}{@{}llrr@{}}
\toprule
Method & Classes & Baseline & + VODA (Mean $\pm\sigma$) \\
\midrule
\multirow{3}{*}{OpenGaussian}
& 19 & 26.20 & $28.32\pm1.12$ \\
& 15 & 28.00 & $28.75\pm1.00$ \\
& 10 & 34.57 & $35.93\pm0.99$ \\
\midrule
\multirow{3}{*}{LaGa}
& 19 & 32.20 & $34.51\pm1.25$ \\
& 15 & 37.19 & $38.26\pm1.06$ \\
& 10 & 43.06 & $45.51\pm1.24$ \\
\bottomrule
\end{tabular}
\end{table}

\begin{table}[htbp]
\centering
\small
\setlength{\tabcolsep}{7pt}
\renewcommand{\arraystretch}{1.10}
\caption{Run-to-run variability on Removal-3D.
Mean $\pm\sigma$ is computed over three VODA runs with fixed baseline results.}
\label{table:repeat_removal}
\begin{tabular}{@{}llrr@{}}
\toprule
Method & Metric & Baseline & + VODA (Mean $\pm\sigma$) \\
\midrule
\multirow{2}{*}{OpenGaussian}
& mIoU $\uparrow$ & 52.87 & $58.16\pm2.62$ \\
& mRDE $\downarrow$ & 39.31 & $30.69\pm2.92$ \\
\midrule
\multirow{2}{*}{LaGa}
& mIoU $\uparrow$ & 75.93 & $77.71\pm1.53$ \\
& mRDE $\downarrow$ & 36.93 & $22.08\pm0.45$ \\
\midrule
\multirow{2}{*}{Gaussian Grouping}
& mIoU $\uparrow$ & 26.23 & $30.23\pm0.68$ \\
& mRDE $\downarrow$ & 50.05 & $36.74\pm2.50$ \\
\midrule
\multirow{2}{*}{ILGS}
& mIoU $\uparrow$ & 31.66 & $37.62\pm0.99$ \\
& mRDE $\downarrow$ & 49.98 & $36.59\pm1.24$ \\
\bottomrule
\end{tabular}
\end{table}

\clearpage
\subsection{Real-World Quantitative and Perceptual Evaluation}
\label{sec:appendix_realworld}

\paragraph{Paired real-world captures.}
We evaluate object removal on SPIn-NeRF~\cite{spinnerf} (10 scenes) and 360-USID~\cite{wu2025aurafusion} (7 scenes), which provide captures of scenes both with and without a target object.
For each scene, we reconstruct the with-object capture, retrieve the target Gaussian set with each baseline or its VODA-enhanced counterpart, remove the selected Gaussians, and render the result.
We evaluate OpenGaussian and Gaussian Grouping, using object-mask IoU on annotated views to assess selection.
For a real-world proxy of RDE, we reconstruct the with-object and without-object captures independently using vanilla 3DGS and use their rendered depth difference as the reference removal signature.
Unlike Removal-3D, this reference is estimated from reconstructions rather than ground-truth geometry.
We also compare removal renderings with the without-object reference images using PSNR, SSIM, and LPIPS.

VODA improves object-mask IoU in all four dataset--baseline settings.
However, the proxy RDE changes are small, with a slight regression for OpenGaussian on 360-USID, and the image-space changes are mixed.
These datasets mainly contain single-object, low-occlusion captures, in which baseline selections already recover most of the depth change.
Moreover, image-space metrics conflate removal completeness with the quality of newly exposed regions, which were not observed during reconstruction.
Thus, these evaluations support transfer to real reconstructions but do not directly establish volumetric completeness under heavy occlusion.

\begin{table}[htbp]
\centering
\small
\setlength{\tabcolsep}{7pt}
\renewcommand{\arraystretch}{1.10}
\caption{Real-world object removal on paired-capture datasets.
Each entry reports baseline $\rightarrow$ $+$VODA.
Object-mask IoU measures selection accuracy, while proxy RDE uses a depth-change reference derived from separate 3DGS reconstructions of the paired captures.}
\label{table:realworld_selection}
\begin{tabular}{@{}llcc@{}}
\toprule
Dataset & Method & Mask IoU $\uparrow$ & Proxy RDE $\downarrow$ \\
\midrule
\multirow{2}{*}{SPIn-NeRF (10)}
  & OpenGaussian & $0.339\rightarrow0.493$ & $9.22\rightarrow9.20$ \\
  & Gaussian Grouping & $0.675\rightarrow0.728$ & $9.82\rightarrow9.78$ \\
\midrule
\multirow{2}{*}{360-USID (7)}
  & OpenGaussian & $0.195\rightarrow0.207$ & $0.601\rightarrow0.609$ \\
  & Gaussian Grouping & $0.347\rightarrow0.553$ & $3.034\rightarrow2.911$ \\
\bottomrule
\end{tabular}
\end{table}

\begin{table}[htbp]
\centering
\footnotesize
\setlength{\tabcolsep}{5pt}
\renewcommand{\arraystretch}{1.10}
\caption{Image-space quality of real-world object removal, compared with the without-object reference captures.
Each entry is baseline $\rightarrow$ $+$VODA.}
\label{table:realworld_image_quality}
\begin{tabular}{@{}llccc@{}}
\toprule
Dataset & Method & PSNR $\uparrow$ & SSIM $\uparrow$ & LPIPS $\downarrow$ \\
\midrule
\multirow{2}{*}{SPIn-NeRF}
  & OpenGaussian & $18.70\rightarrow18.29$ & $0.667\rightarrow0.666$ & $0.329\rightarrow0.331$ \\
  & Gaussian Grouping & $19.79\rightarrow19.82$ & $0.711\rightarrow0.711$ & $0.273\rightarrow0.272$ \\
\midrule
\multirow{2}{*}{360-USID}
  & OpenGaussian & $17.79\rightarrow17.88$ & $0.565\rightarrow0.571$ & $0.316\rightarrow0.313$ \\
  & Gaussian Grouping & $18.56\rightarrow19.08$ & $0.563\rightarrow0.569$ & $0.262\rightarrow0.252$ \\
\bottomrule
\end{tabular}
\end{table}

\paragraph{VLM-based perceptual evaluation.}
We additionally evaluate removal renderings from real LERF scenes using a VLM judge (GPT-5.6).
Baseline and $+$VODA results are scored independently on a five-point quality scale, with method identities hidden and left--right placement balanced.
For each paired comparison, a win indicates a higher score for that method; equal scores are counted as ties.
Table~\ref{table:realworld_perceptual} reports the comparison counts.

\begin{table}[htbp]
\centering
\small
\setlength{\tabcolsep}{6pt}
\renewcommand{\arraystretch}{1.10}
\caption{VLM preferences for real-scene (LERF) removal quality.
$n$ counts evaluated cases; ties are cases with equal scores.}
\label{table:realworld_perceptual}
\begin{tabular}{@{}lrrrr@{}}
\toprule
Method & $n$ & Baseline wins & $+$VODA wins & Ties \\
\midrule
Overall & 80 & 12 & 41 & 27 \\
OpenGaussian & 57 & 12 & 21 & 24 \\
Gaussian Grouping & 23 & 0 & 20 & 3 \\
\bottomrule
\end{tabular}
\end{table}
\FloatBarrier

The VLM judge favors VODA in the number of non-tied comparisons, particularly for Gaussian Grouping.
These judgments provide complementary perceptual evidence on real scenes without requiring paired removal captures; they do not replace the ground-truth depth evaluation on Removal-3D.

\subsection{Qualitative Results and Failure Modes}
\label{sec:appendix_qualitative}

\begin{figure}[htbp]
\centering
\includegraphics[width=0.98\linewidth]{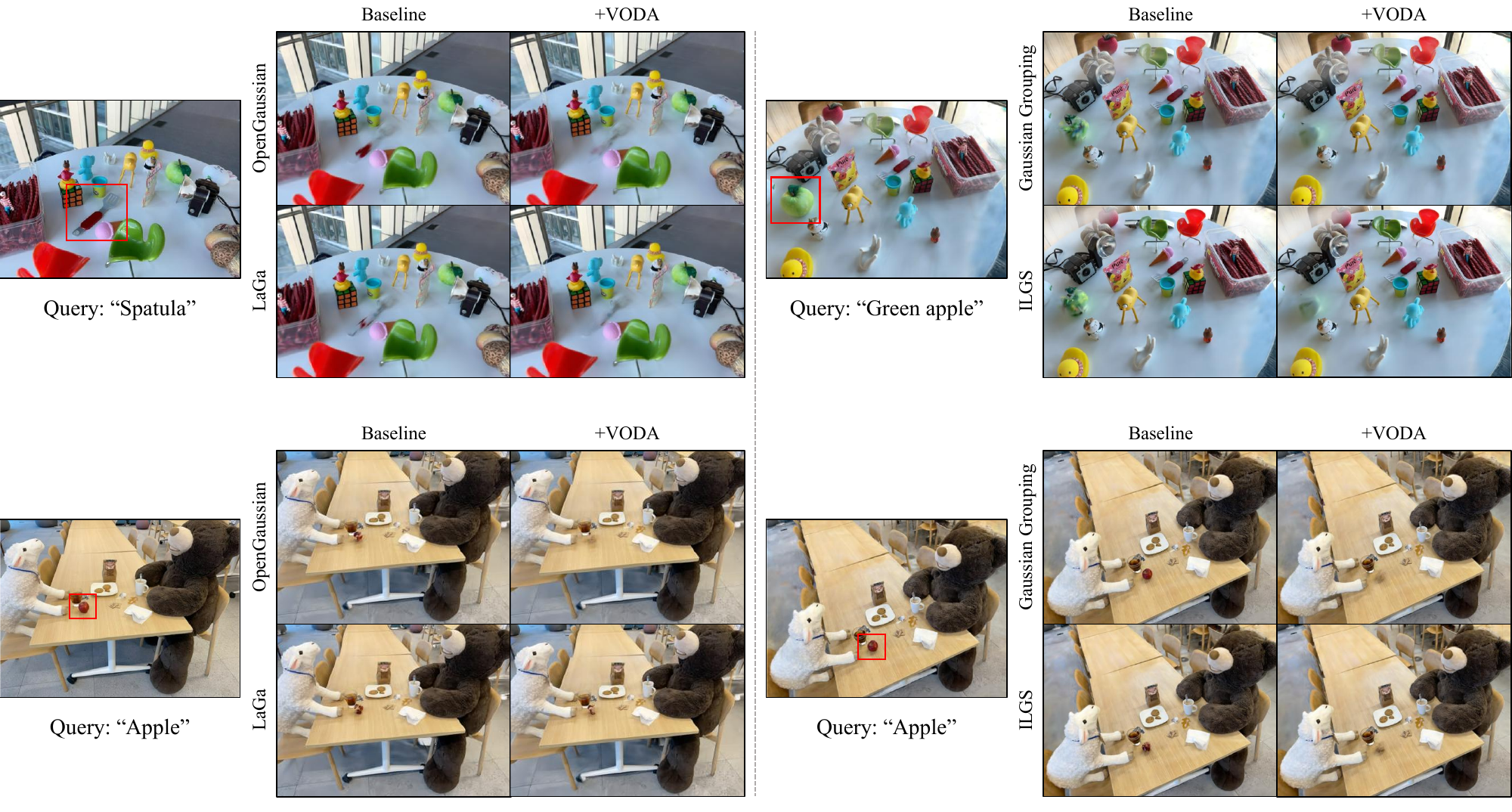}
\caption{Qualitative object selection and Gaussian removal on real LERF scenes.
Red boxes mark the queried objects, and the accompanying renderings show the corresponding scene views.}
\label{fig:lerf_removal}

\end{figure}

\paragraph{Real-world results on LERF.}
To complement the synthetic Removal-3D benchmark, we show qualitative object queries on real LERF scenes in Fig.~\ref{fig:lerf_removal}.
The examples illustrate object selection and scene rendering after Gaussian removal in captured environments.

\paragraph{Failure cases.}
Failure examples highlight two recurring limitations.
Depth clustering can under-split contacting objects with similar depths, while sparse object interiors can leave residual Gaussians after removal.

\begin{figure}[htbp]
\centering
\includegraphics[width=0.98\linewidth]{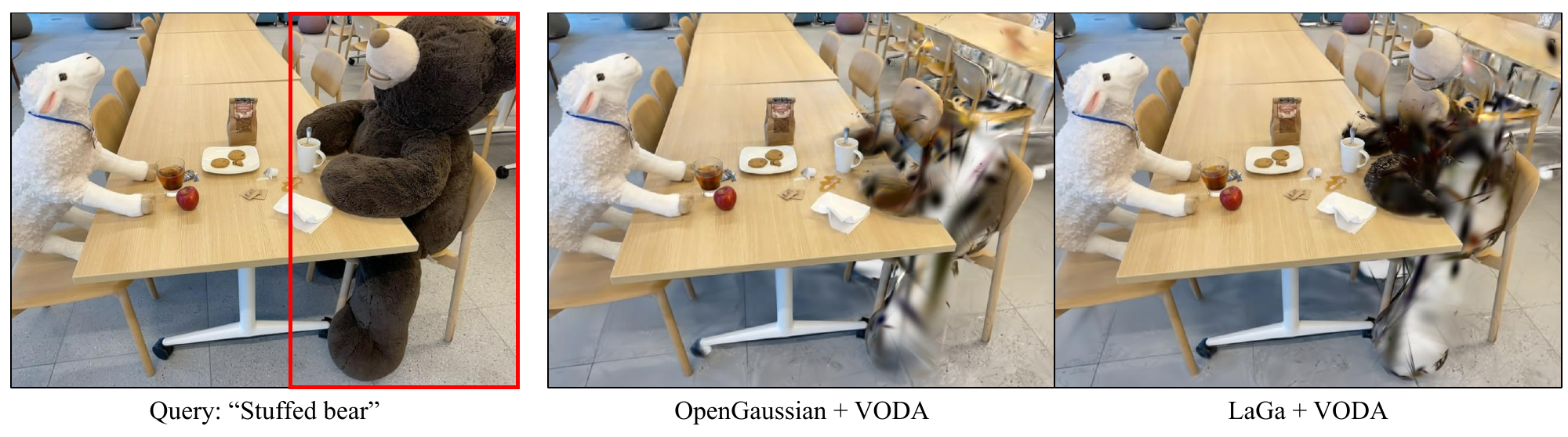}
\caption{Failure case on a real LERF scene.
The queried teddy bear is outlined in red in the input view, and the renderings after Gaussian removal show substantial artifacts around the target region.}
\label{fig:supp_qual_failure}
\end{figure}
\FloatBarrier


\clearpage
\section*{NeurIPS Paper Checklist}

\begin{enumerate}

\item {\bf Claims}
    \item[] Question: Do the main claims made in the abstract and introduction accurately reflect the paper's contributions and scope?
    \item[] Answer: \answerYes{}
    \item[] Justification: The main claims made in the abstract and introduction accurately reflect the
paper’s contributions and scope.

\item {\bf Limitations}
    \item[] Question: Does the paper discuss the limitations of the work performed by the authors?
    \item[] Answer: \answerYes{}
    \item[] Justification: In the end part of our main paper, we discuss the limitations of this paper.

\item {\bf Theory assumptions and proofs}
    \item[] Question: For each theoretical result, does the paper provide the full set of assumptions and a complete (and correct) proof?
    \item[] Answer: \answerNA{}
    \item[] Justification: This paper does not present formal theoretical results requiring proofs.

    \item {\bf Experimental result reproducibility}
    \item[] Question: Does the paper fully disclose all the information needed to reproduce the main experimental results of the paper to the extent that it affects the main claims and/or conclusions of the paper (regardless of whether the code and data are provided or not)?
    \item[] Answer: \answerYes{}
    \item[] Justification: We will provide open-source codes.

\item {\bf Open access to data and code}
    \item[] Question: Does the paper provide open access to the data and code, with sufficient instructions to faithfully reproduce the main experimental results, as described in supplemental material?
    \item[] Answer: \answerNA{}
    \item[] Justification: We will provide it later.

\item {\bf Experimental setting/details}
    \item[] Question: Does the paper specify all the training and test details (e.g., data splits, hyperparameters, how they were chosen, type of optimizer) necessary to understand the results?
    \item[] Answer: \answerYes{}
    \item[] Justification: The details of training and testing are provided in A.1.

\item {\bf Experiment statistical significance}
    \item[] Question: Does the paper report error bars suitably and correctly defined or other appropriate information about the statistical significance of the experiments?
    \item[] Answer: \answerYes{}
    \item[] Justification: We write them in Appendix.

\item {\bf Experiments compute resources}
    \item[] Question: For each experiment, does the paper provide sufficient information on the computer resources (type of compute workers, memory, time of execution) needed to reproduce the experiments?
    \item[] Answer: \answerYes{}
    \item[] Justification: We write them in Appendix.

\item {\bf Code of ethics}
    \item[] Question: Does the research conducted in the paper conform, in every respect, with the NeurIPS Code of Ethics \url{https://neurips.cc/public/EthicsGuidelines}?
    \item[] Answer: \answerYes{}
    \item[] Justification: The research conducted in the paper conforms, in every respect, with the
NeurIPS Code of Ethics.

\item {\bf Broader impacts}
    \item[] Question: Does the paper discuss both potential positive societal impacts and negative societal impacts of the work performed?
    \item[] Answer: \answerNA{}
    \item[] Justification: There is no societal impact of the work performed.

\item {\bf Safeguards}
    \item[] Question: Does the paper describe safeguards that have been put in place for responsible release of data or models that have a high risk for misuse (e.g., pre-trained language models, image generators, or scraped datasets)?
    \item[] Answer: \answerNA{}
    \item[] Justification: Not related.

\item {\bf Licenses for existing assets}
    \item[] Question: Are the creators or original owners of assets (e.g., code, data, models), used in the paper, properly credited and are the license and terms of use explicitly mentioned and properly respected?
    \item[] Answer: \answerYes{}
    \item[] Justification: The models and data used in this paper are open-sourced and authorized.

\item {\bf New assets}
    \item[] Question: Are new assets introduced in the paper well documented and is the documentation provided alongside the assets?
    \item[] Answer: \answerNA{}
    \item[] Justification: The paper does not release new assets.

\item {\bf Crowdsourcing and research with human subjects}
    \item[] Question: For crowdsourcing experiments and research with human subjects, does the paper include the full text of instructions given to participants and screenshots, if applicable, as well as details about compensation (if any)?
    \item[] Answer: \answerNA{}
    \item[] Justification: No human-participant evaluation is reported in the final manuscript.

\item {\bf Institutional review board (IRB) approvals or equivalent for research with human subjects}
    \item[] Question: Does the paper describe potential risks incurred by study participants, whether such risks were disclosed to the subjects, and whether Institutional Review Board (IRB) approvals (or an equivalent approval/review based on the requirements of your country or institution) were obtained?
    \item[] Answer: \answerNA{}
    \item[] Justification: No human-participant evaluation is reported in the final manuscript.


\item {\bf Declaration of LLM usage}
    \item[] Question: Does the paper describe the usage of LLMs if it is an important, original, or non-standard component of the core methods in this research? Note that if the LLM is used only for writing, editing, or formatting purposes and does \emph{not} impact the core methodology, scientific rigor, or originality of the research, declaration is not required.
    \item[] Answer: \answerYes{}
    \item[] Justification: To polish writing.

\end{enumerate}

\end{document}